\documentclass[9.5pt,journal,compsoc]{IEEEtran}
\usepackage{amsmath,amsfonts,amssymb}
\usepackage{algorithmic}
\usepackage{algorithm}
\usepackage{array}
\usepackage[caption=false,font=normalsize,labelfont=sf,textfont=sf]{subfig}
\usepackage{textcomp}
\usepackage{stfloats}
\usepackage{url}
\usepackage{verbatim}
\usepackage{graphicx}
\usepackage{cite}
\usepackage{xcolor}
\usepackage{epsfig}
\usepackage{booktabs}
\usepackage{makecell}
\usepackage{multirow}
\usepackage{pifont}
\newcommand{\cmark}{\ding{51}}%
\newcommand{\xmark}{\ding{55}}%
\usepackage{tablefootnote}
\usepackage[bookmarks=false,colorlinks]{hyperref}

\usepackage{enumitem}

\usepackage{xcolor}

\begin{document}
\IEEEtitleabstractindextext{
\begin{abstract}

Recent artificial intelligence (AI) technologies show remarkable evolution in various academic fields and industries. However, in the real world, dynamic data lead to principal challenges for deploying AI models. An unexpected data change brings about severe performance degradation in AI models. We identify two major related research fields, {\it domain shift} and {\it concept drift} according to the setting of the data change. Although these two popular research fields aim to solve distribution shift and non-stationary data stream problems, the underlying properties remain similar which also encourages similar technical approaches.
In this review, we regroup domain shift and concept drift into a single research problem, namely the data change problem, with a systematic overview of state-of-the-art methods in the two research fields. We propose a three-phase problem categorization scheme to link the key ideas in the two technical fields. We thus provide a novel scope for researchers to explore contemporary technical strategies, learn industrial applications, and identify future directions for addressing data change challenges.
\end{abstract}

}

\title{A Comprehensive Review of Machine Learning Advances on Data Change: A Cross-Field Perspective}


\author{Jeng-Lin Li,~\IEEEmembership{Member,~IEEE,}
        Chih-Fan Hsu,~\IEEEmembership{Member,~IEEE,}
        Ming-Ching Chang,~\IEEEmembership{Senior Member,~IEEE,}
        Wei-Chao Chen,~\IEEEmembership{Senior Member,~IEEE,}
\thanks{Jeng-Lin Li, Chih-Fan Hsu, and Wei-Chao Chen are with Inventec Corporation, Taipei 111, Taiwan. EMail: \url{li.johncl@inventec.com}, \url{hsuchihfan@gmail.com}, \url{chen.wei-chao@inventec.com}.}
\thanks{Ming-Ching Chang is with University at Albany -- State University of New York, Albany, NY 12065, USA. EMail: \url{mchang2@albany.edu}.}
\thanks{Manuscript received April XX, 2023; revised August XX, 2023. \\
Corresponding authors: Jeng-Lin Li, Chih-Fan Hsu, Ming-Ching Chang, and Wei-Chao Chen}
}

\markboth{Journal of \LaTeX\ Class Files,~Vol.~XX, No.~X, XXX~2023.}%
{Shell \MakeLowercase{\textit{et al.}}: A Sample Article Using IEEEtran.cls for IEEE Journals}

\IEEEpubid{0000--0000/00\$00.00~\copyright~2021 IEEE}

\maketitle
\IEEEdisplaynontitleabstractindextext

\section{Introduction}

\IEEEPARstart{D}{ata}-centric Machine Learning (ML) paradigms have emerged into important technical approaches to address many challenges in various Artificial Intelligence (AI) application fields, including the automotive industry~\cite{alvarez2021towards}, internet-of-things (IoT)~\cite{8246999}, and healthcare~\cite{jarrahi2022principles}. 
Data is pivotal in driving model learning, influencing performance, and serving as an infrastructure in real-world systems.
However, data may change due to differences in the sources, unexpected sensor corruption, and seasonal variations. New data distributions might gradually deviate from the original distributions. As a result, the predictive capability of the learned model can degrade accordingly. In the literature, there are two scenarios widely investigated with respect to data changes: {\bf domain shift}~\cite{zhang2022transfer,jiang2022transferability,tan2018survey,kouw2018introduction} and {\bf concept drift}~\cite{hu2020no,wares2019data,8571222,8496795,9544693,hashmani2020concept}. The frameworks for domain shift aim to handle data changes caused by the change of data sources with the unchanged model's predicted performance. In comparison, concept drift methods aim to handle data changes caused by time-varying phenomena leading to the obsolescence of existing models.
Although the original problem definitions differ in the two scenarios, recent ML advancements in the two fields share commonalities and trends, due to the similar underlying properties of {\bf data change}.

\begin{figure}[t]
\centerline{
  \includegraphics[width=\linewidth]{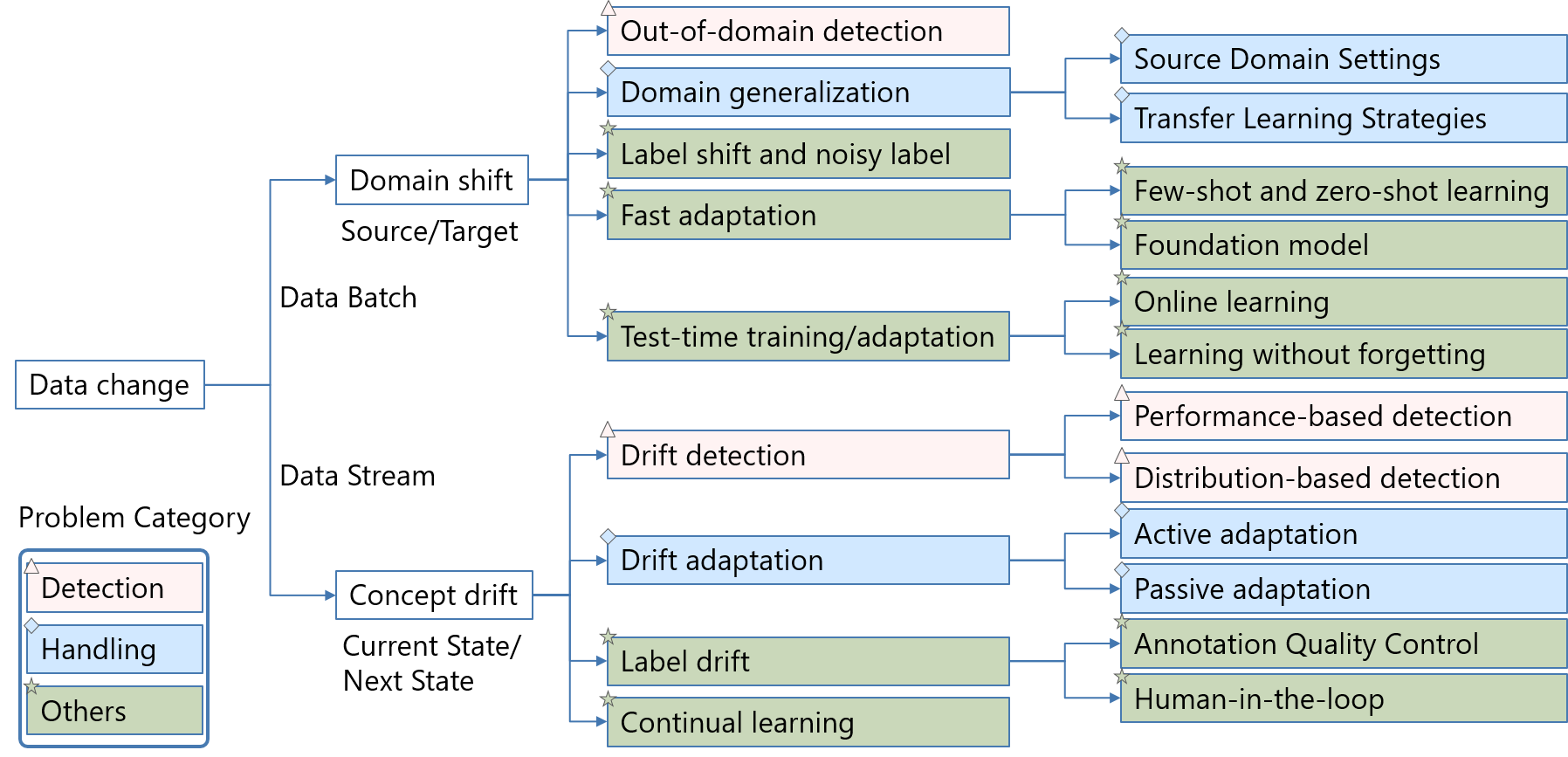}
\vspace{-3mm}
}  
\caption{
  {\bf Taxonomy of the data change corresponds to this paper's sections and sub-sections.} We illustrate our three-phase problem categorization scheme in different colors and icons. The topics of domain shift and concept drift are categorized into these phases.}
\label{fig:taxonomy}
\end{figure}

\begin{figure*}[t]
\centerline{
  \includegraphics[width=0.95\linewidth]{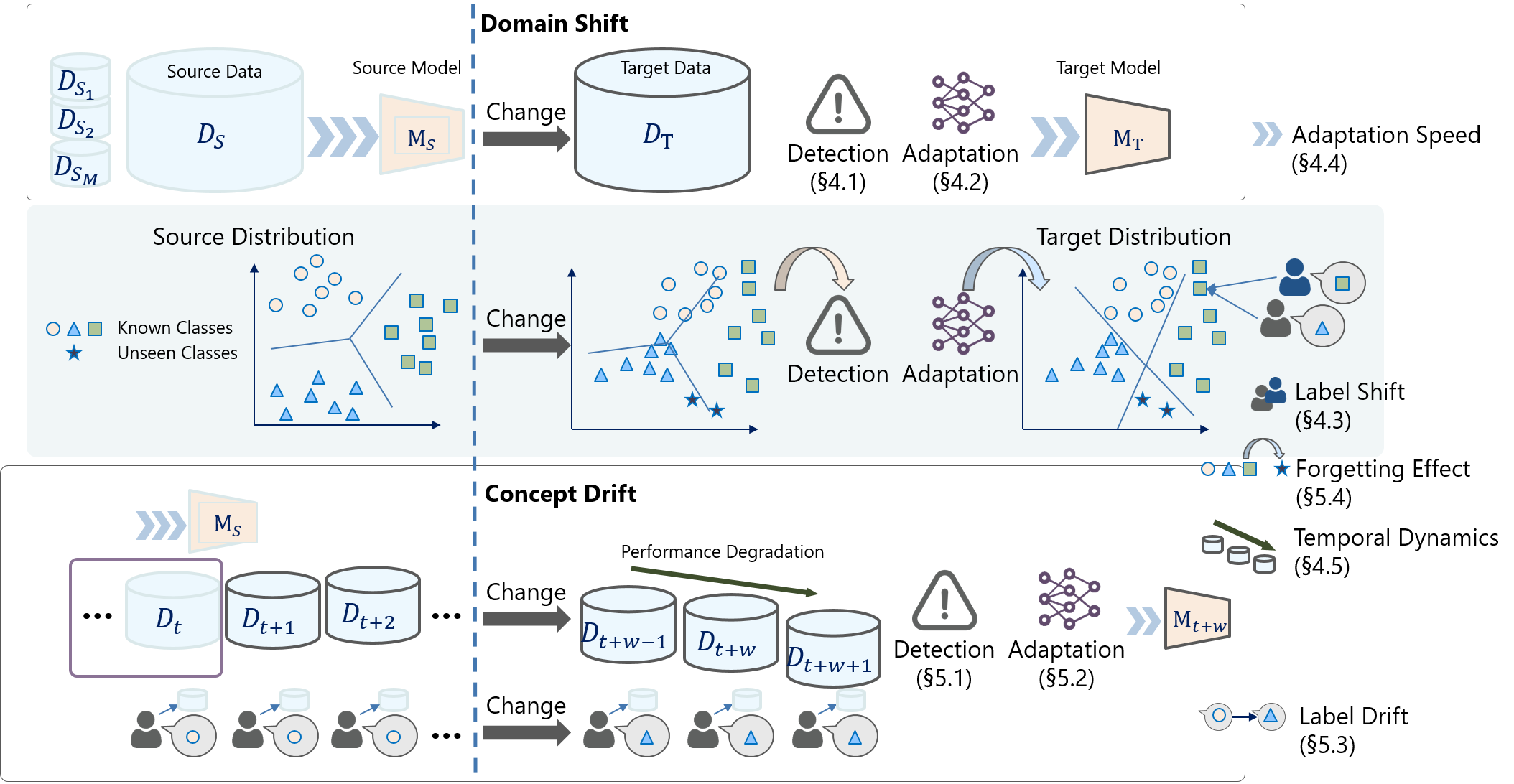}
\vspace{-3mm}
}  
\caption{
{\bf Overview of research topics related to the data change problems.} The corresponding section is written in parentheses brackets followed by the research problem. The source domain model $M_S$ is trained with the data $D_S$ and the target domain model and data are denoted as $M_T$ and $D_T$. Data from $D_t$ to $D_{t+w}$ denote the data in an observed window for concept drift. Refer to $\S$\ref{sec:notation} for the corresponding notations and relations.}
\label{fig:framework}
\end{figure*}

Domain adaptation aims to bridge the gaps between the source and target domains~\cite{farahani2021brief}. The problem intricately encompasses distribution shifts in both data and labels, arising from changes in data quality, annotating criteria, and the targeted label classes. 
Substantial research endeavors have been dedicated to enhancing the generalization capability of models in various scenarios, leveraging the concept of distribution learning~\cite{ganin2016domain,Xu_Zhang_Ni_Li_Wang_Tian_Zhang_2020,yu2020label}.
A common strategy is to train a robust and general source model and then fine-tune it for downstream tasks.

Concept drift arises from the need to monitor model performance throughout the lifecycle of a system. Unexpected drift often occurs in data streams in non-stationary environments, where data distributions evolve over time, potentially leading to catastrophic failures in real-world applications. A common strategy to address concept drift is estimating temporal changes by analyzing historical data and adapting the model to accommodate forthcoming changes.

Regarding the underlying data change, concept drift represents a form of temporal distribution shift that requires timely adaptation. Therefore, it is natural to align with the concept of domain shift. Researchers have expanded and begun to explore similar approaches to handle the two problems.
For example, continual learning, which is specifically designed to mitigate {\em catastrophic forgetting}, can help address domain shift problems towards universal model learning. Simultaneously, continual learning techniques were also developed to incrementally learn new knowledge without forgetting important historical knowledge from the data stream~\cite{lesort2021understanding} in the context of concept drift. Similarly, regularization and ensemble learning techniques are frequently leveraged in both domains.
Fast domain adaptation and sequential test-time training are two sub-fields within the domain shift methods that emphasize efficiency and temporal factors during learning. Interestingly, these factors align with the primary challenges addressed in concept drift. Although the original assumptions in the two fields have shaped distinct algorithmic framework developments, emerging topics have led to an intertwining of technical approaches across these fields. 

In this paper, we propose a unified perspective to bridge the literature gap between domain shift and concept drift.
We further provide insightful industrial perspectives on the challenges associated with this gap~\cite{9124702} for real-world model deployment applications. 
This paper aims to answer the following research questions (RQs):
\begin{itemize}[leftmargin=10pt] \itemsep -.1em
    \item (RQ1) What problem settings have been proposed to tackle and consolidate the complicated data change issues in the research fields?
    \item (RQ2) What are the current state-of-the-art approaches that address the problems under these settings?
    \item (RQ3) How best to contribute to the integration of research fields revolving around the data change problems in a unified categorization scheme?
    \item (RQ4) What are the perspectives, challenges, and practical applications in industries associated with data change?
\end{itemize}
\noindent

Based on their underlying shared data change characteristics, we associate domain shift and concept drift with a new and unified {\bf three-phase scheme} to make evident the advantages of techniques in the two fields, facilitating comparisons of framework design ideas across the fields.
Specifically, Fig.~\ref{fig:taxonomy} shows the three-phase scheme including problem detection, problem handling, and other related factors. These phases suggest the logic of the model deployment procedure starting from detecting the problem to adapting models and finally incorporating other factors to further relax usage constraints. Efficiency is an important factor in adaptation, as it helps prevent the burden of extensive model retraining. Continual learning for the forgetting effects has also been studied to achieve sustainable model deployment in concept drift. Fig.~\ref{fig:taxonomy} lists detailed terms used in the fields of domain shift and concept drift. For instance, in the case of domain shift, data undergo changes in batches, whereas, in concept drift, data changes occur in streams. Consequently, in domain shift, the source and target domains represent the data before and after the change, whereas in concept drift, the current and next states are utilized to capture the evolving data dynamics.



Overall, this paper shows a comprehensive overview of the global landscape, cutting-edge designs, and emerging trends of domain shift and concept drift. Specifically, we summarize the key research and surveys in both research fields and introduce a problem-oriented taxonomy to categorize the state-of-the-art studies. We examine the major difference and similarities between the methods and datasets, thereby identifying potential avenues for future research. Additionally, we highlight other important topics, such as model bias and fairness, as well as advances in the related communities that should be considered in addressing data change problems.
We expect this overview paper can stimulate technical improvement to address the challenging data change problems and guide researchers and practitioners for future development.

Our contributions are summarized in the following. 
\begin{itemize}[leftmargin=10pt] \itemsep -.1em
    \item Our work bridges the gaps of key ideas for handling data change in modern deep learning that spans multiple research fields around domain shift and concept drift. 
    \item This study reviews the state-of-the-art methods and illustrates the characteristics of the two fields, thereby bringing researchers to the frontiers of cutting-edge advancements.
    \item We introduce a three-phase scheme including problem identification, problem handling, and extended factors, to categorize the research topics and illuminate key developments and emerging topics in the two fields.
    \item We propose future research directions regarding the challenges of model deployment in industrial applications.
\end{itemize}

The rest of the paper is organized as follows: $\S$\ref{sec:overview} summarizes the recent survey papers on domain shift and concept drift, and $\S$\ref{sec:notation} describes the notations used in this paper. $\S$\ref{sec:domain adaptation} and $\S$\ref{sec:concept drift} introduce the domain shift and the concept drift problems with a series of research topics using our proposed problem categorization scheme. $\S$\ref{sec:discussion} discusses the shared and complementary techniques in the fields of domain shift and concept drift. $\S$\ref{sec:industrial_perspective} provides an industrial perspectives for the state-of-the-art studies. Finally, $\S$\ref{sec:conclusion} concludes this data change study with the current status and future trends.


\begin{table}[t]
\addtolength{\tabcolsep}{-5pt}
\caption{Recent survey papers published in 2018-2023.}
\label{table:related_work}
\vspace{-2mm}
\begin{tabular}{ll}
\toprule
Domain Shift              & Concept Drift        \\ \midrule
Transfer Learning~\cite{zhang2022transfer,jiang2022transferability, tan2018survey} & General~\cite{8571222,8496795,hashmani2020concept}  \\
Unsupervised Adaptation~\cite{8861136,SIP-2022-0019,9238468}  & Unsupervised Learning~\cite{9289802,Fahy2022,AGRAHARI20229523}\\  
Source-free Adaptation~\cite{yu2023comprehensive} & Machine learning ~\cite{SUAREZCETRULO2023118934,jameel2020critical} \\ 
Domain Generalization~\cite{zhou2022domain} & Regression~\cite{9762269} \\ 
Vision~\cite{SurveyMin2022,9238468} & Class imbalance~\cite{8246564}  \\ 
Language~\cite{saunders2022domain,guo2022domain,chan2023state} & Deep learning~\cite{ijcai2022p788} \\ 
Acoustic and Speech~\cite{yadav2022survey,9296327}   & Performance-based approach~\cite{BAYRAM2022108632}         \\
Reinforcement Learning  \cite{zhu2020transfer}    & Distance measurement~\cite{mahmood2021concept}\\
\bottomrule
\end{tabular}\vspace{-2mm}
\end{table}

\section{Related Surveys and Datasets}
\label{sec:overview}


Table~\ref{table:related_work} lists the existing survey papers up to 2023 that are on domain shift and concept drift. We briefly introduce their key contributions and the featured topics to summarize the surveys in both fields. Table~\ref{tab:domain_dataset} and Table~\ref{tab:drift_dataset} summarize the commonly-used public datasets for the evaluation of domain shift and concept drift, respectively.  

\subsection{Domain Shift Survey Papers}
\label{ssec:related_domain_adaptation}

Domain shift involves the transfer learning across domains in addressing the data distribution shift problems. With the identified source and target domains, the domain shift problem can easily be found when we need to deal with multiple datasets. Therefore, the techniques are extensively investigated in multiple prominent machine learning applications, such as vision, language, and speech. Table~\ref{table:related_work} lists the recent survey papers along with the key topics that shows the trend of domain adaptation development. The general focus on transfer learning has garnered significant attention, with notable contributions from four survey studies~\cite{zhang2022transfer,jiang2022transferability,tan2018survey,kouw2018introduction}. Additionally, unsupervised domain adaptation has emerged as a widely discussed topic~\cite{8861136,SIP-2022-0019,9238468}. Vision~\cite{SurveyMin2022,9238468}, language~\cite{saunders2022domain,guo2022domain,chan2023state}, and speech~\cite{yadav2022survey,9296327} applications have also been organized to lighlight the technologies developed in the contexts, each with its own task and data properties. Review studies have also covered topic such as reinforcement learning~\cite{zhu2020transfer}, providing a comprehensive understanding of the research findings.


\begin{table*}[t]
\addtolength{\tabcolsep}{1pt}
\caption{{\bf Existing benchmark datasets for domain shift.} We describe the major data change factors along with the modality of the datasets.
\vspace{-3mm}
}
\label{tab:domain_dataset}
\centerline{
\begin{tabular}{c|c|c|c|c|c}
\toprule
\multicolumn{5}{c}{\textbf{Vision}} & \\
\midrule
Attribute  & Data Corruption & Noisy Data  & Style     & Environment  & \\
\midrule
\makecell{Colored MNIST~\cite{arjovsky2019invariant}, \\ MNIST-R~\cite{ghifary2015domain}} & \makecell{CIFAR10-C~\cite{hendrycks2019benchmarking},\\ CIFAR100-C~\cite{hendrycks2019benchmarking},\\ ImageNet-C~\cite{hendrycks2019benchmarking},\\ ModelNet40-C~\cite{sun2022benchmarking},\\ VisDA-C~\cite{visda2017}}  & \makecell{ANIMAL-10N~\cite{pmlr19b_animaln},\\ CIFAR-10N~\cite{wei2022learning},\\ CIFAR-100N~\cite{wei2022learning}, \\Clothing1M~\cite{xiao2015learning},\\ WebVision~\cite{li2017webvision}} & \makecell{PACS \cite{li2017deeper},\\ Office-Home~\cite{venkateswara2017deep},\\ DomainNet \cite{zhao2019multi},\\ ImageNet-R~\cite{hendrycks2021many},\\ VisDA~\cite{visda2017}} & \makecell{VLCS~\cite{fang2013unbiased},\\ Wilds~\cite{koh2021wilds},\\ NICO~\cite{he2021towards}} \\ 
\midrule
\midrule
\multicolumn{4}{c|}{\textbf{Language}} & \multicolumn{2}{c}{\textbf{Speech}}  \\
\midrule
Language (Translation) & Query & Dialog Intention  & Sentiment & Data Source  & Language \\
\midrule
\makecell{CoCoA-MT~\cite{nǎdejde2022cocoa}} & \makecell{OSQ~\cite{9052492}} & \makecell{SNIPS~\cite{coucke2018snips},\\CLINC150~\cite{larson-etal-2019-evaluation},\\ MultiWOZ 2.2~\cite{zang-etal-2020-multiwoz}} 
& \makecell{
ACE2005~\cite{walker2006ace}, \\ Amazon MDSD~\cite{blitzer-etal-2007-biographies}} & \makecell{LibriAdapt~\cite{9053074}} & \makecell{Common voice~\cite{commonvoice:2020}}  \\
\bottomrule
\end{tabular}
}
\end{table*}

\subsection{Domain Shift Datasets}
\label{ssec:domain_shift_datasets}

We investigate the publicly available datasets commonly used in domain shift research.
Existing datasets are organized into three setup types according to their covering domains, the use of synthetic data, and the data sources.
First, to cover broad domain changes, samples were collected in-the-wide from multiple domains~\cite{karouzos-etal-2021-udalm,chu2022denoised}. 
Second, data change can be simulated by changing the color and viewing angle, or by adding noise or blurriness~\cite{arjovsky2019invariant,ghifary2015domain,hendrycks2021many} and other data corruption~\cite{hendrycks2019benchmarking,sun2022benchmarking,visda2017}.  
Third, a simple approach to prepare a cross-domain dataset is to combine two or more naturally different datasets collected from multiple sources or institutions~\cite{Ao_Li_Ling_2017,ganin2016domain,8273627,lu2018learning}. 




Table~\ref{tab:domain_dataset} provides an overview of datasets collected with domain shift data in different modalities.
Sub-fields of domain shift such as {\em out-of-domain detection} ($\S$\ref{ssec:OOD}) and {\em domain generalization} ($\S$\ref{ssec:domain generalization}) are developed on large-scale benchmark datasets to include different types of data change. The commonly considered domains are style and environmental changes, such as clipart and infographic styles in DomainNet~\cite{zhao2019multi}, and speaker accents and acoustic environment in LibriAdapt~\cite{9053074}. Changes in language tasks naturally involve cross-lingual contents~\cite{walker2006ace} and conversation topics~\cite{zang-etal-2020-multiwoz}. 
These studies frequently use cross-institution setups while turning into the multi-domain setup as the scale of data collection grows.
Fast adaptation ($\S$\ref{ssec:fast adaptation}) follows a similar data configuration, yet it selects the dataset altered with distinct label classes.
Studies on label shift ($\S$\ref{ssec:label_shift}) and test-time training and adaptation ($\S$\ref{ssec:ttt tta}) often utilize synthesized corrupted data ({\em e.g.}, CIFAR100-C or ImageNet-C~\cite{hendrycks2019benchmarking}) to represent the target domain with presumed change types.


\subsection{Concept Drift Survey Papers}
\label{ssec:related_concept_drift}

Numerous studies have investigated the issue of concept drift that occurs in various machine-learning applications. Common categories of algorithms include performance-based detection~\cite{BAYRAM2022108632}, distribution-based detection~\cite{mahmood2021concept}, hypothesis-based detection, and contextual-based detection~\cite{BAYRAM2022108632}. The majority of previous review studies have identified similar taxonomies for categorizing typical concept drift approaches~\cite{hu2020no,wares2019data,8571222,8496795,9544693,hashmani2020concept}. Table~\ref{table:related_work} lists the topics highlighted in the review papers between 2018 and 2023. Unsupervised approaches are widely used due to their ability to effectively handle concept drift without relying on labeled data. Fast-growing machine learning techniques are then employed to address concept drift~\cite{SUAREZCETRULO2023118934,jameel2020critical}. Meta-learning has emerged as a promising direction for concept drift adaptation~\cite{SUAREZCETRULO2023118934}. 
Algorithms tailored to specific usage scenarios, such as regression~\cite{9762269} and class imbalance~\cite{8246564} problems, are developed taking into account the properties of the respective problems. Furthermore, studies in deep learning as presented in~\cite{ijcai2022p788}, have focused on learning network parameters and architectures to address concept drift in the data stream. 



\begin{table*}[t]
\addtolength{\tabcolsep}{-4pt}
\caption{
{\bf Commonly used datasets for the concept drift.} 
We describe the major data change factors along with the task descriptions.
\vspace{-3mm}
}
\label{tab:drift_dataset}
\centerline{
\begin{tabular}{l|l|l}
\toprule
Dataset & Prediction Goal & Descriptions (Change Factor) \\
\midrule
Weather~\cite{5975223} & raining prediction  & climate change over 1949-1999 \\
Electricity~\cite{zliobaite2013good} & daily price changes (rise or fall) & change of prices affected by supply and demand \\
Forest~\cite{blackard1999comparative} & seven cover types & change of ecological processes  \\
Outdoor Objects~\cite{7280610} & forty objects &  change of lighting conditions on a mobile in a garden environment \\
Rialto Bridge Timelapse~\cite{7837853} & ten colorful buildings & change of weather and lighting conditions over 20 consecutive days \\
Luxembourg~\cite{vzliobaite2011combining} & frequency of internet usage (high or low) & change of personal habits and political tendency \\
Chess~\cite{vzliobaite2011combining} & outcome of the game (won, lost, or draw) & change of player skills and opponent selection by system \\
\bottomrule
\end{tabular}
}
\end{table*}

\subsection{Concept Drift Datasets}
\label{ssec:concept_drift_datasets}

Current studies on concept drift have used both real and synthetic datasets, with a focus on continuous temporal data. Tabular and structured data are more commonly employed in these cases. Given the diverse conditions under which concept drift occurs in different data collection scenarios, we report commonly used real datasets in Table~\ref{tab:drift_dataset}. In line with the trend of utilizing real-world data with an unbiased estimation of the drift, we exclude synthetic datasets and prioritize the frequently used datasets in the table.
The selected datasets comprise the prediction targets that are influenced by various potential drift factors. For example, the weather observations exhibit varying frequencies of rainy days as a consequence of climate change~\cite{5975223}. Similarly, ecological changes impact forest coverage~\cite{blackard1999comparative}. Video recordings often exhibit changes in lighting conditions and recording environments~\cite{7280610,7837853}. Human behaviors and habits also exhibit temporal variations, which can be observed in internet usage patterns~\cite{vzliobaite2011combining} and chess playing~\cite{vzliobaite2011combining}.
For research aiming at comprehensive studies across various applications, Souza {\em et al.} have presented additional benchmark datasets~\cite{souza2020challenges}.

\begin{table}[t]
\caption{{\bf Commonly referred terms} in domain shift and concept drift.
\vspace{-3mm}
}
\label{tab:notation}
\centerline{
\begin{tabular}{ccc}
\toprule
              & Domain Shift         & Concept Drift         \\
\midrule
Data Access   & Batch                & Stream                \\
Original Data & Source Domain ($D_S$) & Previous State ($D_t$) \\
Changed Data  & Target Domain ($D_T$) & Next State ($D_{t+w}$)  \\
\bottomrule
\end{tabular}
}
\end{table}

\begin{table*}[t]
\caption{
{\bf Problem definition and properties for the data change technical fields.} $D_m$ indicates the $m$-th domain dataset used for model training.
\vspace{-3mm}
}
\label{tab:problem_definition}
\centerline{
\begin{tabular}{l|cccc|cccc}
\toprule
\multirow{2}{*}{Domain Shift} & \multicolumn{4}{c|}{Training}  & \multicolumn{4}{c}{Change}   \\
\cmidrule(lr){2-9}
& \makecell{Source Data}    & \makecell{Source Model}   & \makecell{Target Data}   & \makecell{Target Model}   & \makecell{Task $\mathcal{Y}$}   & \makecell{$P(Y)$}   & \makecell{$P(Y|X)$} & \makecell{Temporal}   \\
\midrule
Ouf-of-Domain Detection       & $D_1$, ... , $D_M$ & $D_1$, ... , $D_M$ & $D_{M+1}$, Batch & \xmark & \xmark & \checkmark & \checkmark & \xmark \\
Domain Generalization         & $D_1$, ... , $D_M$   & $D_1$, ... , $D_M$ & $D_{M+1}$, Batch & \xmark & \checkmark\footnotemark/\xmark & \xmark & \checkmark & \xmark  \\
Domain Adaptation             & $D_i$ & $D_i$ & $D_j$, Batch & $D_j$ & \checkmark\footnotemark/\xmark & \xmark & \checkmark & \xmark \\
Label Shift                   & $D_i$ & $D_i$ & $D_j$, Batch & $D_j$ & \xmark & \checkmark & \checkmark & \xmark  \\
Fast Adaptation               & $D_1$, ... , $D_M$ & $D_1$, ... , $D_M$ & $D_{M+1}$, Batch & $D_{M+1}$ & \checkmark & \xmark & \checkmark\footnotemark/\xmark & \xmark \\
Test-Time Training            & \xmark & $D_i$ & $D_j$, Stream & $D_j$ & \xmark & \xmark & \checkmark & Short \\
\midrule
\multirow{2}{*}{Concept Drift} & \multicolumn{4}{c|}{Training}  & \multicolumn{4}{c}{Change}   \\
\cmidrule(lr){2-9}
& \makecell{Source Data}    & \makecell{Source Model}   & \makecell{Target Data}   & \makecell{Target Training}   & \makecell{Task $\mathcal{Y}$}   & \makecell{$P(Y)$}   & \makecell{$P(Y|X)$} & \makecell{Temporal}   \\
\midrule
Drift Detection               & $D_i$ & $D_i$ & $D_j$, Stream & \xmark & \xmark & \checkmark/\xmark & \checkmark/\xmark & Long \\
Drift Adaptation              & $D_i$ & $D_i$ & $D_j$, Stream & $D_j$ & \xmark & \checkmark/\xmark & \checkmark/\xmark & Long \\
Label Drift                   & $D_i$ & $D_i$ & $D_j$, Stream & $D_j$ & \xmark & \checkmark & \xmark & Long \\
Continual Learning            & $D_1$, ... , $D_M$ & $D_1$, ... , $D_M$ & $D_j$, Stream &  $D_j$  & \checkmark & \xmark & \xmark & Short    \\
\bottomrule
\end{tabular}
}
\vspace{2mm}
\footnotemark[1]{Zero-shot learning}  \hspace{4mm}
\footnotemark[2]{Incremental learning}  \hspace{4mm}
\footnotemark[3]{Cross-domain few-shot learning}\vspace{-3mm}
\end{table*}


\section{Notations and Definitions}
\label{sec:notation}

We next introduce the notations used in this paper. We aim to establish consistent notations and definitions to describe the data change associated with domain shift and concept drift for all papers that are reviewed. A specific domain is characterized by a distribution $P(X)$, which represents a set of data samples $X=\{x_1, x_2, ..., x_N\}$. Here, $N$ denotes the number of samples in the collected dataset $D$. For a given machine learning task $\mathcal{T}$, which involves a collections of labels $Y=\{y_1, y_2, ..., y_N\}$ in the dataset $D$, supervised learning can be conducted using the labeled samples ${x_i, y_i}$ for $i\in N$, $x_i\in X$ and $y_i\in Y$. 


We generalize the notations in the context of domain shift to describe the data change for concept drift. Specifically, we define the mismatch between a source domain and a target domain for domain shift and further extend this definition to address concept drift. This allows the dynamic updating of the two domains to accommodate the data stream setting. Table~\ref{tab:notation} shows the unified definition for the various terms commonly used in the fields of domain shift and concept drift. The original data correspond to the source domain dataset $D_S$ and the changed data correspond to the target domain dataset $D_T$. Concept drift is usually defined on a particular time step $t$ and a time window $w$, to specify the data in the current state $D_t$ and the occurrence of drift data at $t+w$, denoted as $D_{t+w}$. Here, $D_t$ can be regarded as a special case of $D_S$, where the time stamp is specified, and thus $D_{t+w}$ can be represented by $D_T$. 

The data change problem is formulated as the distribution shift between $P_S(X, Y)$ and $P_T(X, Y)$ which contains the special case of the concept drift problem.
Table~\ref{tab:problem_definition} provides an overview of the sub-field problems addressed in this paper. We highlight the differences in data and models between the source and target domains to illuminate the changes. As there might be multiple source or target domain datasets, we denote the dataset as $D_m$, where $m$ represents the index of the domain. The checklist showing the data change of the distribution is also presented in Table~\ref{tab:problem_definition}.


\section{Domain Shift}
\label{sec:domain adaptation}

We categorize the research efforts addressing the distribution shift between the source domain distribution $P_S(X, Y)$ and target domain distribution $P_T(X, Y)$ in three phases of problem categorization, as shown in Fig.~\ref{fig:taxonomy}. The {\em problem detection} phase focuses on out-of-domain detection technologies ($\S$\ref{ssec:OOD}). The {\em problem handling} phase encompasses domain adaptation and generalization approaches ($\S$\ref{ssec:domain generalization}). Additional factors including label quality, adaptation speed, and temporal dynamics are considered in $\S$\ref{ssec:label_shift}, $\S$\ref{ssec:fast adaptation}, and $\S$\ref{ssec:ttt tta}.   

\subsection{Out-of-domain Detection}
\label{ssec:OOD}

Out-of-domain (OOD) detection is vital in real-world applications, as it can avoid data abnormality that deviates from the typical distribution. This topic was recently identified in 2017~\cite{yang2021oodsurvey} due to the limitations of common domain adaptation approaches in ensuring safe and robust deployment in rapidly changing environments. From the perspective of domain adaptation, the OOD detection problem aims to identify unknown data sample that differs from the source domain dataset $D_S$, which is typically unavailable during the test phase. $D_S$ is referred to as in-domain data and the unknown samples are OOD data. The goal is to identify samples that do not belong to the source distribution $P_S(X, Y)$. Therefore, the OOD problem can seen as an early-phase monitoring problem, as the target dataset $D_T$ is not accessible for retraining. 
The scope of OOD detection has expanded to include various related topics such as anomaly detection, novelty detection, open set recognition, out-of-distribution detection, and outlier detection~\cite{yang2021oodsurvey}. This broader perspective considers not only changes in sample features but also unknown classes and noises within the domain shift context. Differentiating between relevant information and irrelevant noises is a complex task, as the definition of in-domain distribution must be unbiased. Overconfident predictions can undermine model trust.


Classification-based methods are intuitive in thresholding the prediction score to distinguish in-domain and out-of-domain distributions. Advanced methods extend the scoring functions to the gradient space, offering better flexibility for observing abnormal patterns. The design of the scoring function, by estimating the probabilistic properties of true likelihood distribution, can keep sensitive scores for model performance~\cite{Morteza_Li_2022}.   
State-of-the-art OOD detection approaches favor agnostic solutions that are applicable to any network structure. For instance, the Simple Activated Function designed for deep networks makes no assumptions on the model architectures~\cite{NEURIPS2021_01894d6f}.
Another effective approach is the non-parametric nearest-neighbor distance, which is distribution assumption-free, test data agnostic, and model agnostic properties~\cite{pmlrOOD2022}. The success of the distance measurement approach can be attributed to techniques like contrastive learning~\cite{winkens2020contrastive} and self-supervised learning~\cite{Mohseni_Pitale_Yadawa_Wang_2020}, which come with enhanced generalization ability for in-domain training.


OOD detection in visual applications aims to detect domain shift factors, such as spatial location accuracy, appearance diversity, image quality, and aspect distribution~\cite{du2022siren}. Speech and language applications may encounter challenges related to cross-language problems or content differences influenced by cultural or contextual factors~\cite{9052492,9741317}. In a recent survey study~\cite{yangopenood}, open source codes are released to facilitate comprehensive benchmarking of the generalized OOD detection frameworks. Other libraries are also available to accelerate the R\&D in this area~\cite{Kirchheim_2022_CVPR}.
These OOD detection approaches have led to the development of corresponding strategies in machine learning systems in two directions: {\em learning with rejection} and {\em domain adaptation}. Rejecting prediction on unknown data maintains the quality of the services and engaging humans to rectify the data promotes effective collaboration between humans and machines for further model learning.
The direction of adaptation described in $\S$\ref{ssec:domain generalization} aims for a more automatic process to adapt models as soon as OOD is identified.

\subsection{From Domain Adaptation to Domain Generalization}
\label{ssec:domain generalization}

Traditional domain adaptation methods assume that we have access to the target domain dataset $D_T$ to estimate the distribution $P_T(X, Y)$, which usually restricts real-world applications. Therefore, unsupervised domain adaptation has emerged, leaving to the promising field of domain generalization (DG). DG encompasses an ambitious goal to generalize learning without accessing to the target domain dataset $D_T$. It can be seen as a more generalized problem of domain adaptation, which aims to handle unseen domains without retraining models~\cite{zhou2022domain}. Both domain generalization and domain adaptation are designed in the problem-handling phase which has attracted abundant research. In contrast to the OOD problem, the DG problem focuses on detecting changes in data across different domain distributions and adapting models to maintain predictive ability even on unseen classes~\cite{wang2022generalizing}. 

\subsubsection{Source Domain Settings}
\label{sssec:source_setting}

The source domain data for training yields in-depth impacts on the model generalization ability and transferability. Herein, we regard two source domain training data settings, including multiple-source and single-source settings.
The typical DG studies use multiple training datasets for multiple sources and attempt to generalize the source domain distribution $P_S$~\cite{zhou2022domain}. The single source DG is more difficult because there is less data variability for the source model. Some prior works deem the single source DG as an OOD generalization problem~\cite{Qiao_2020_CVPR} to focus on the samples' learning once the OOD has been detected.
The typical domain adaptation studies use the single source setting that assumes the availability of the target domain labels for adaptation. 

Aside from the two data source settings for training, source-free domain adaptation prohibits the source data to be used for adaptation once the source model has been trained~\cite{Wang_2022_CVPR,9200758}.
The adaptation is only based on the trained model parameters without access to the source domain data $D_S$. Technically, clustering along with pseudo labels can help estimate the distribution shift~\cite{Ding_2022_CVPR}. Xie et al. propose contrastive category matching to exploit the relationship between pairs of targeted images~\cite{Xia_2021_ICCV}. Source-free domain generalization has been identified as a harsh problem. Although teacher-student learning~\cite{frikha2021towards} and using text prompts as a proxy can be possible solutions~\cite{niu2022domain}, there still remains substantial space for further research.

\subsubsection{Transfer Learning Strategies}
\label{sssec:transer}

Research in both domain adaptation and domain generalization has grown with three branches of learning strategies to conquer the domain gap problem including domain finetuning, domain adversarial learning, and domain matching approaches. Their distinct underlying ideas shape the framework designs with more focus on model refining, distribution difference elimination, and distribution alignment techniques, respectively.

\textbf{Domain Finetuning:} The domain finetuning is the most fundamental approach for domain adaptation which can then be introduced to domain generalization with additional algorithm adjustment. 
Researchers are meant to construct a large and robust source model for initialization and design fine-tuning techniques for downstream tasks.
The knowledge distillation approaches extract the knowledge in the source domain network for further reuse in the target domain~\cite{Ao_Li_Ling_2017}.
Teacher-student learning enhances the knowledge distillation approaches by keeping using the distilled teacher network to supervise the finetuning steps~\cite{li2017large}. With the knowledge distillation technique, the frameworks can deal with source-free cases~\cite{frikha2021towards}.
Self-supervised learning loss acts as a regularization term to map the same-class samples across domains close to each other~\cite{kim2021selfreg}.

\textbf{Domain Adversarial Learning:} The domain adversarial learning is a mainstream technique that is renowned for Domain Adversarial Neural Network (DANN) introducing the reverse gradient with an additional discriminator to avoid one domain being differed from another~\cite{ganin2016domain}. This setting naturally results in an adversarial condition in which one branch learns to recognize the main task and the other branch disables the ability to discriminate different domains. The adversarial approaches further improve the convergence by using mixup to create continuous label space instead of hard domain labels~\cite{Xu_Zhang_Ni_Li_Wang_Tian_Zhang_2020}. Although adversarial learning is designed to learn domain invariant latent features, over-optimization on domain discriminator can degrade the recognition performance on the target domain. Accordingly, explicitly designing an interplay of the classifier and the domain discriminator attains better adversarial optimization~\cite{Tang_Jia_2020}. Recent visual transformer (ViT) takes advantage of the attention module to re-weight image patches to boost transferability during the adversarial learning~\cite{Yang_2023_WACV}. The adversarial learning can also be extended to domain generalization studies by seeking a common ground of multiple datasets in the domain discriminator~\cite{gideon2019improving}.

\textbf{Domain Matching:} The domain matching approaches minimize the domain discrepancy by optimizing feature consistency across domains. The focus of the research turns to the design of distribution measurement approaches, such as maximum mean discrepancy (MMD)~\cite{rozantsev2018beyond}, correlation alignment~\cite{sun2016return}, contrastive domain discrepancy~\cite{Kang_2019_CVPR}, and Wasserstein distance~\cite{shen2018wasserstein}. 
Other than the distribution measurement approaches, batch normalization can adaptively dissociate bias and variance of the datasets~\cite{li2016revisiting}. 
In the source-free scenario, measuring classifier discrepancy with pseudo-labeled samples facilitates self-training~\cite{chu2022denoised}.

\subsection{Label Shift and Noisy Labels}
\label{ssec:label_shift}

Most machine learning models highly rely on labeled data for supervised learning. However, unreliable labels might be observed in situations of label shift and noisy labels. Label shift refers to the change of the underlying label distribution $P_S(Y)\neq P_T(Y)$. Although noisy labels would also give rise to the difference between the label distribution. We differ the label shift and noisy labels with the corresponding learning strategies that mainly keep the target distribution or the source distribution characteristics, respectively. 

Current studies blindly estimate the label shift problem and performance drop as it occurs unexpectedly. For example, Lipton et al. have proposed a black box shift estimation on an invertible confusion matrix and the distribution ratio before and after label shift~\cite{lipton2018detecting}. To generalize the estimation in the low target sample setting, an importance weighting approach with regularization to address three kinds of synthesized label shift~\cite{azizzadenesheli2019regularized}. 
These studies quantify the shift with mathematical forms and propose a statistical bound to guarantee the bound of generalization. However, the presumed labeled shift might be oversimplified. An advanced work proposes a model to jointly estimate sparse covariate shift and label shift~\cite{chen2022unsupervised}. Recently, online learning algorithms can be used to actively estimate the shift~\cite{NEURIPS2021_5e6bd7a6}.

One of the situations of label shift is that the source domain is contaminated with noisy labels. 
Data annotation engages in the cognitive status and past experiences of the annotators which yields potential labeling errors when the task is complex or the annotators are biased. The noisy labels in the source domain impede the model's robustness to generalize across domains. These corrupted labels might occupy 8.0\% to 38.5\% in real-world datasets~\cite{song2022learning}. The label quality issues are amplified as crowd-sourced labeling solutions can reduce data annotation costs. 
Technically, learning with noisy labels aims to handle the problem by enhancing the robust source domain model learning~\cite{song2022learning,kumar22_interspeech} including framework designs on architecture, regularization, loss design, and sample selection. Curriculum learning is a remedy to recover noisy data as clean data by the estimation of empirical risk~\cite{han2021towards}.
One of the loss designs is the denoising MMD loss to estimate the target domain label distribution with noisy source data~\cite{yu2020label}.
Sample selection by measuring the divergence of two classifiers for each sample can alleviate the adverse effect of noisy samples and perform partial alignment on the cross-domain distributions~\cite{Yu_2021_CVPR}.

A special case related to the learning from noisy labels is the source-free domain adaptation in that recent studies have generated pseudo labels as noisy labels for the target data using a pre-trained source model~\cite{9981099,yisource}. Regularizing the source model to avoid overly adapting to the noisy target data provide a bound of effect from the noise~\cite{yisource}. Separating target domain samples into clean and noisy data also regularizes the learning in a self-supervised manner~\cite{9981099}. Considering the label shift as noise has opened up new avenues for robust learning across domains.

\subsection{Fast Adaptation}
\label{ssec:fast adaptation}

Frequent change in real-world data poses challenges to the domain adaptation models with the need for data efficiency and convenience. This section includes algorithms resolving the issues with inconsistent label space and target domain tasks. 
Few-shot learning and zero-shot learning are notable solutions to deal with learning with limited access to new data. A well-trained source domain model is the key to generalizing without complicated operations on the target domain data. Sophisticated techniques are proposed to improve this source domain model training to derive the so-called foundation model. The idea is that the foundation model aggregating massive knowledge provides strong initialization for any tasks to transfer with minimal effort.
To allow fast adaptation with limited testing data, the technical approaches are optimized on either pretraining or finetuning phases. The pretraining aims to learn a generic model with large-scale datasets and the finetuning requires several techniques to efficiently adapt model parameters.

\subsubsection{Few-Shot Learning}
\label{sssec:fewshot learning}

Few-shot learning is a problem set upon a harsh condition that the target dataset $D_T$ contains only very few available labeled samples for adaptation. We formally define a task $\mathcal{T}_S=\{\mathcal{Y}, P(Y|X)\}$ with the label space $\mathcal{Y}$ and conditional distribution $P(Y|X)$. 
The common setup includes the source domain dataset $D_S$ for typical supervised training and validation. Then, a new $C$-way $K$-shot task $\mathcal{T}'$ is set up in the situation that $\mathcal{T}\neq \mathcal{T}'$. That is, there are $C$ unseen classes in the target domain, resulting in disjoint label space.
However, only a very limited number of $K$ samples are available as the target domain support set $D_{T}^{S}$ for model training, and the final task is to predict on a query set $D_{T}^{Q}$. 
In the context of a typical pre-trained model utilizing dataset $D_S$, the occurrence of domain shift towards dataset $D_T$ poses a challenge known as cross-domain few-shot learning. This shift is characterized by the dissimilarity between the probability distributions $P_S(X)$ and $P_T(X)$.


A prior survey has categorized few-shot learning research into meta-learning and non-meta-learning methods~\cite{parnami2022learning}. 
The task change naturally leads the few-shot learning algorithm to seek more tasks for pretraining. Meta-learning becomes a major approach, mimicking the prediction of new tasks with sampling strategies (e.g., MAML~\cite{Chi_2022_CVPR}). 
Data augmentation and prototypical networks are notable examples of non-meta-learning methods~\cite{survey_fewshot2023}. The design principle of these approaches is to expand the feature space with sufficient sample diversity, providing a robust semantic structure for the new task to finetune delicate information. 
The cross-domain few-shot learning problem is more challenging with the change of both the domain and task. Associating prototypical network and adversarial learning serves as a way to capture new task prototypes in the new domain~\cite{zhao2021domain}. Moreover, training with a contrastive loss enhances the semantic structure of the self-supervised prototypes~\cite{yue2021prototypical}. 
Domain-Switching Learning embeds the knowledge from multiple domains to realize a fast switch to the target domain~\cite{hu2022switch}. The setting is similar to domain generalization but allows a few data in the support set to tackle the task differences. Zero-shot learning is an extended field restricting the use of the support set while introducing semantic features (attributes) to bridge to the new tasks~\cite{pourpanah2022review}.

\subsubsection{Foundation Models}
\label{sssec:foundation model}

The progress of the pre-trained model with continuously enlarged source domain data size has achieved promising success in the natural language processing~\cite{zhou2023comprehensive}, vision~\cite{ijcai2022p762,wang2023internimage}, and audio~\cite{baevski2020wav2vec} domains for fast adaptation to various downstream tasks. The core idea of the foundation model is to aggregate as much knowledge as possible into a pre-trained model for generic use. 

\textbf{Pretrained Model: }
Large-scale self-supervised learning is a dominated source-domain model training approach with the idea that predicts the latent features of masked inputs with self-training. Contrastive learning can minimize the distance between positive pairs of samples and maximize negative pairs of samples while the non-contrastive approach only considers the positive pairs~\cite{tian2021understanding}. Intriguingly, the recent research indicates that the covariance regularization-based non-contrastive methods can be regarded as performing contrastive pairs on dimensions rather than samples~\cite{garrido2023on}. The learning methods for pre-trained models might share similar optimization goals with different techniques. In addition, Data2vec demonstrates that the same self-supervised training approach can be applied to different modalities~\cite{baevski2022data2vec}. The foundation model can not only support multimodal tasks~\cite{ijcai2022p762,radford2021learning}, but also encompass huge information with even cross-domain samples. For example, cross-lingual acoustic and language models have been introduced by training with multiple source domain datasets~\cite{babu2021xlsr,scao2022bloom}.

\textbf{Finetuning: }
With a pre-trained foundation model, parameter efficiency in finetuning becomes crutial that requires minimal effort for deployment in new tasks, which is also termed as ``delta tuning''~\cite{ding2022delta}. Typically, researchers train an adapter by inserting a small learnable module to avoid updating the whole bulky network~\cite{pmlr_houlsby19a}. Prefix tuning manipulates the keys and values attention heads~\cite{li-liang-2021-prefix} and prompt tuning simplifies the manipulation to the input tokens~\cite{lester2021power}. Instead of inserting an adapter module, researchers also introduce a learnable low-rank matrix to modify the projection of intermediate layers~\cite{hu2021lora}. A unified view has been proposed to cast the problem as a learning problem on modification vectors of hidden layers~\cite{he2022towards}.

\subsection{Test-time Training and Adaptation}
\label{ssec:ttt tta}

One of the emerging topics of unsupervised domain adaptation, test-time training (TTT) or test-time adaptation (TTA), has been consolidated recently that the generalization of learning in the testing phase can leverage unlabeled testing data as an indicator for updating the model on the new data distribution~\cite{sun19ttt}. This setting breaks the traditional stationary assumption and indicates that the testing data distribution might change over time. In practice, we observe a small batch of testing data and update the model in an unsupervised manner. Then, we evaluate the performance on-the-fly with the adapted model.   
TTA can be thought of as an online domain adaptation problem combining the techniques from domain generalization and fast adaptation described in $\S$\ref{ssec:domain generalization} and $\S$\ref{ssec:fast adaptation}, respectively. During training, researchers design the learning of source domain models for generalization while during testing, self-training with the unlabeled test data at hand needs to adjust the model for a better fit. In this task, the adaptation at test time is usually conducted without access to the source domain data. This setting connects TTA with source-free DG described in $\S$\ref{ssec:domain generalization}.

\subsubsection{Online Learning}
\label{sssec:online_learning}

TTT depends on the powerful foundations of source domain training and the strategies of the target domain unsupervised adaptation. The designed loss function with self-supervised learning techniques dominates the success of TTT. In addition, meta-learning is another strategy to expand the source domain learning space for a generic model, and thus the model simultaneously exploits the learned knowledge from other tasks~\cite{Chi_2021_CVPR}. 
In the testing phase, computing the self-supervised learning loss on the unlabeled testing data can update the model to handle new data distribution. The distribution mismatch often causes failure of adaptation due to the missing link across domains.   
Therefore, selecting an appropriate loss and matching the samples located in a similar feature space are techniques to diminish the distribution shift. For example, Goyal et al. examine training losses’ convex conjugate to facilitate to seek for the best TTA loss for a given source model and distribution shift~\cite{goyaltest}. Chen et al. refine online pseudo labels with soft voting among their nearest neighbors and learn to contrast positive and negative samples~\cite{chen2022contrastive}. 
Liu et al. propose a feature alignment approach by computing statistics to represent the distribution characteristics \cite{NEURIPS2021_b618c321}.  
Gandelsman et al. reframe the TTT problem as one-sample learning which aims to perfectly fit each testing sample point for tailored prediction~\cite{gandelsmantest}.
Sequential test-time training (sTTT) is recently defined to strictly include only one-pass adaptation on the data stream~\cite{su2022revisiting}. That is, refining the model with the accumulated testing data is not allowed to minimize computational overhead. Even further, adjustment on the classifier completely avoids the back-propagation and minimizes the cost at inference time~\cite{iwasawa2021test}.

\subsubsection{Learning without Forgetting}
\label{sssec:TTA_continual}

The model updating over time raises the catastrophic forgetting issue which learns the pattern from the new data samples but loses the discriminative power on the old samples. The new pattern is usually attributed to the out-of-domain samples described in $\S$\ref{ssec:OOD}. Continual learning, also known as lifelong learning, has been exploited as a remedy to consolidate the learned knowledge over time.  
For example, Niu et al. propose to actively select reliable samples with regularized updating to avoid dramatic network parameters change~\cite{niu2022efficient}. 
Qin et al. reduce the error accumulation by using weight-averaged and augmentation-averaged prediction and introducing stochastic restoring partial neurons during training iterations to avoid the forgetting effects~\cite{wang2022continual}. 
The model updating efficiency is also emphasized in these studies because the frequent back-propagation during the testing phase leads to great burdens on computational consumption. Samples redundantly generating similar gradients can be discarded to ensure sample efficiency during adaptation~\cite{niu2022efficient}. The studies imply the data-efficient learning trend for data stream with minimal updating and maximal knowledge consolidation.

\section{Concept Drift}
\label{sec:concept drift}

For the input of data streams, the ability for models to adapt on the fly is crucial when facing dynamic changes in data distribution. Typical machine learning models are ill-suited for dealing with non-stationary scenarios, which significantly impacts their reliable deployment across diverse application domains. 
In the studies~\cite{BAYRAM2022108632,8496795}, the term ``concept drift'' is defined as
{\em the occurrence of a change in the joint probability of variables $X$ and $Y$ at time $t$, within a specified time window $w$}.
Specifically, $\exists t, w: P_t(X, Y) \neq P_{t+w}(X, Y)$. To maintain consistency, we refer to the distribution at time $t$ as the source distribution and the distribution at time $t+w$ as the target distribution. Consequently, the concept drift is expressed in a similar manner as the domain shift problem, where $P_S(X, Y) \neq P_T(X, Y)$, based on the notation defined in $\S$\ref{sec:notation}.

We identify three distinct types of potential sources of drift from the literature survey. 
The first type is referred to as {\em virtual draft}, characterized by the condition where $P_S(X) \neq P_T(X)$ while $P_S(Y|X)=P_T(Y|X)$. This form of drift does not explicitly impact the decision boundary.
The second type, known as {\em actual drift}, occurs when $P_S(Y|X) \neq P_T(Y|X)$ while $P_S(X)=P_T(X)$. In this case, there is a direct influence on the posterior probability for predicting the target label, thus affecting the decision boundary.
Lastly, the third type, termed {\em prior probability drift}, arises when $P_S(Y) \neq P_T(Y)$. This drift significantly alters the learning process, often involving novel classes or the disappearance of existing classes.


%

\begin{figure}[t]
\centerline{
  \includegraphics[width=\linewidth]{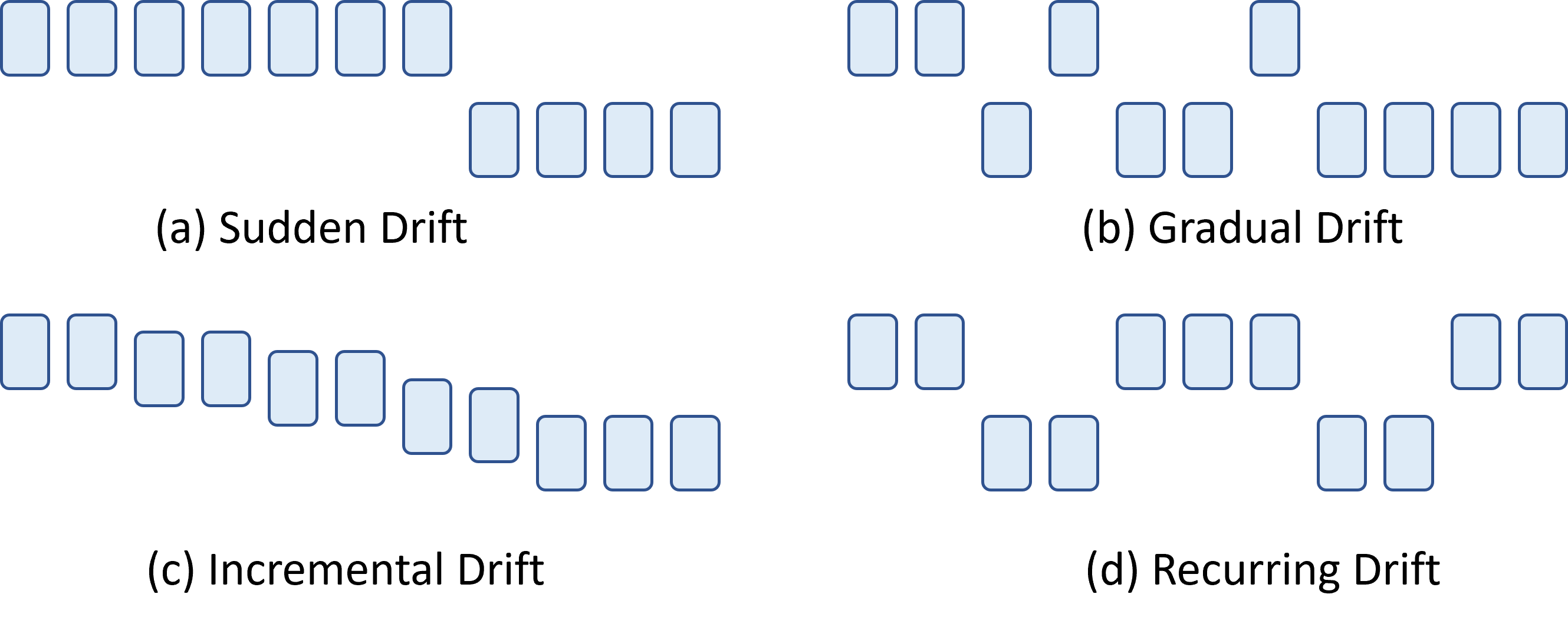}
\vspace{-3mm}
}  
\caption{
{\bf Examples of the common drift types.} 
The figure is adapted from the review paper~\cite{BAYRAM2022108632}.}
\label{fig:drift_type}
\end{figure}

Research topics on concept drift generally concern the why, how, where, and when the drift happens~\cite{8496795}. Factors behind the observed drift can be intricate. For example, social trends involving newly-created terms and preferences can influence text recognition tasks in recommendation systems. Transient, seasonal, and circumstantial changes are sometimes subject to domain-specific and user differences, such as policy changes, pandemics, and personality.
Most existing studies categorize the drifts into the following types~\cite{8496795} as in Fig.~\ref{fig:drift_type}: {\em sudden drift, gradual drift, incremental drift, and recurring concepts}. The general implementation setting is accumulating historical data, training models, and performing testing while the demand for collecting historical data leads to a trade-off between usability and model stability. 
Prior studies are inclined to neglect the concept drift issues in common settings. 
Recent studies have focused on the importance of temporal generalization, arguing that the random split used in the previous works might unintentionally neglect the issue of concept drift, which hinders their real-life usage~\cite{sogaard-etal-2021-need}.

According to the proposed three-phase problem categorization in Fig.~\ref{fig:taxonomy}, we investigate the drift detection ($\S$\ref{ssec:drift detection}) and drift adaptation ($\S$\ref{ssec:drift_adaptation}) for the first two phases. We discuss other factors including label drift ($\S$\ref{ssec:label drift}) and continual learning ($\S$\ref{ssec:continual learning}), which offer solutions for unstable human annotations and the forgetting effects, respectively.

\subsection{Drift Detection}
\label{ssec:drift detection}

We organize the technical developments in drift detection into three main categories, namely, {\em performance-based}, {\em distribution-based}, and {\em meta-learning-based} methods. Performance-based and distribution-based methods have long development history, and meta-learning is an emerging technique using machine learning algorithms to handle concept drift. 

\subsubsection{Performance-based Detection}
\label{sssec:performance-based detection}

The most intuitive means to detect drift is monitoring the performance change with the assumption that the drift directly results in the degradation of the model accuracy. Bayram {\em et al.} have provided an in-depth review of this kind of performance-based detectors~\cite{BAYRAM2022108632}. The {\em rule-based}, {\em statistical-based}, and {\em ensemble-based} methods are widely used due to the easiness of deployment. Traditional methods usually set thresholds and rules to capture the pattern of different types of drift. For example, monitoring performance drop within an adaptive time window ({\em e.g.}, DDM~\cite{gama2004learning} and ADWIN~\cite{bifet2007learning}) to detect the drift by selecting appropriate rule parameters. Recent advances focus on selecting the proper threshold as an index instead of developing new algorithms~\cite{9729470}. Statistical-based methods formulate the rules into multiple hypothesis tests to alleviate false alarms from insignificant data change cues. Hierarchical Hypothesis Testing was designed upon Heoffding’s inequality and two Kolmogorov-Smirnov (KS) tests to monitor the defined class uncertainty~\cite{yu2018request}. 
Ensemble-based methods detect drafts by examining the performance degradation of multiple learners. One example is to adaptively maintain different levels of ensemble diversity, in a way that ensembles of both high and low diversity can mix detecting patterns to reduce  bias~\cite{sidhu2015online}.

Ensemble-based methods can be combined with rule-based and statistical-based methods. The EnsembleEDIST2~\cite{khamassi2019new} uses EDIST2, a statistical hypothesis test on distribution distance in the defined windows, for decision criteria update. 
Meta-learning is then built based on the multiple-learner setting in the ensemble-based methods. The meta-learner can reuse or replace the base learners by tracking their behaviors~\cite{anderson2019recurring}. 
Recently, Yu {\em et al.}~\cite{YU2022996} pre-trained an offline prototypical network as a meta-detector with meta-features extraction based on error rates; they conducted online finetuning for improved prediction accuracy.

\subsubsection{Distribution-based Detection}
\label{sssec:distribution-based detection}

Distribution-based detection approaches utilize distance functions to measure the difference between the distribution of historical and new data. This measurement is usually examined via hypothesis tests to ensure the significance of the detected results. The underlying concept is similar to {\em anomaly detection} which can detect the distribution variation in an unsupervised manner. 
Under this circumstance, studying clustering algorithms along with a family of distance functions can properly quantify the distribution relations, where the representation space is shaped by the distance function. Zubaroğlu and Atalay surveyed~\cite{zubarouglu2021data} the clustering approaches for data stream with detailed characteristics regarding the time window, computational complexity, and accuracy. 

Centroids as the cluster centers can represent the average characteristics of a subset of data in a distribution, which evolves with the techniques to learn ``prototypes''. The window of focus becomes a parameter for measuring bias from the prototypes. For example, regional drifts can be observed within a larger window. The well-known Gaussian Mixture Model (GMM) allows the measurements to be maintained as soft clusters over time~\cite{oliveira2021tackling}.  
The nearest neighbor-based density variation identification (NN-IDV)~\cite{liu2018accumulating} employs a k-Nearest Neighbor (kNN) model with distance measurements, and statistical tests are used to specify the region with drift occurrence. Other studies attempt to alleviate the unsatisfactory results of unsupervised detection. The supervised embedding training, optimized through the incorporation of intra-class and inter-class relationships with cluster centroids, enables the prediction of real drift by means of distance comparison~\cite{castellani2021task}.
Recent advances have tied in with deep learning. Learning a generative model based on the learned historical latent distribution can detect possible drift clues~\cite{li2022ddg}. The data generator can adapt fast by forecasting the drift before it happens rather than detecting the drift that is already happened. 




\subsection{Drift Adaptation}
\label{ssec:drift_adaptation}

The drift adaptation can be conducted either actively or passively. {\em Active adaptation} ($\S$\ref{sssec:active_adapt}) explicitly performs detection on the data stream and adapts to the new data once detects the drift. {\em Passive adaptation} ($\S$\ref{sssec:passive_adapt}) methods instead update the model continuously to ensure predictive ability without degradation. As for the model updating scheme, both active and passive learning approaches have adopted either partial updating, full updating, or ensemble updating. Note that passive adaptation approaches are tightly associated with the development of partial updating because frequent updates might increase computational overhead.

\subsubsection{Active Adaptation}
\label{sssec:active_adapt}

These approaches heavily rely on the drift detection approaches described in $\S$\ref{ssec:drift detection}. Right after the drift is detected, most studies focus on model retraining straightforwardly. 
Vector quantization approaches are commonly used to generate prototypes which can then be adapted with the parameters after drift detection~\cite{RAAB2020340}.
A recent study in \cite{9748034} moved toward temporal deep learning, by combining the Long Short-Term Memory (LSTM) with an ensemble learning strategy, to estimate a drift factor and assist the adaptation of short-text classification.
The main issue of the active adaptation method is that false alarms might trigger inefficient adaptation. Identifying different types of drift might be a direction to customize the algorithm for the characteristics of the drift. For example, Liu {\em et al.}~\cite{9047166} proposed a diverse instance-weighting ensemble approach for regional drift, which resulted in an excessive number of false alarms during adaptation. 

\subsubsection{Passive Adaptation}
\label{sssec:passive_adapt}

These methods often involve the utilization of evolution algorithms, which gradually adapt the model to ensure robustness in new environments~\cite{ghomeshi2019eacd}. 
Automatic adjusting of the learned parameters of the clusters using stochastic gradient descent techniques can enhance model capability against variation~\cite{heusinger2022passive}.
The design of network modules that can automatically determine adapted parameters is advantageous for modern deep networks.
For instance, in~\cite{9783029}, hybrid spiking neurons embedded in LSTM can provide an attention mechanism for multivariate time series. This approach avoids the requirement of pre-set variance into the standard surrogate gradient.
Automatic parameter adjustment can also be achieved by carefully designing the loss function. In~\cite{chalkidis2022improved}, the group optimization algorithm regularizes the disparity between temporally-split groups. This approach is inspired by group disparity among different attributes of data ({\em e.g.}, gender and age).
Due to the limited accessed data in a time window, switching the adaptation between the minority and majority classes is essential for the class imbalance issue~\cite{li2020incremental}.

Passive adaptation approaches usually accommodate active adaptation schemes through updates triggered by an additional drift detector. Passive adaptation approaches usually come with more complex issues such as adaptation efficiency and class imbalance problems.

\subsection{Label Drift}
\label{ssec:label drift}

While previous research on concept drift has predominantly focused on changes in $P(X|Y)$, the drift of labels where $P(Y)$ changes, has received relatively less attention. Moreover, there are certain factors not adequately addressed in previous concept drift studies, such as manual labeling drift. This drift is intricately connected to the properties of the task at hand and the cognitive states of the annotators involved.
Therefore, {\em annotation quality control} is crucial to effectively detect instances where the annotators' performance declines ($\S$\ref{sssec:robust_training}). Other research further proactively involves {\em human in the loop} ($\S$\ref{sssec:human_in_the_loop}) to foster workflows in real-world scenarios.




\subsubsection{Annotation Quality Control}
\label{sssec:robust_training}


Robustly detecting and adapting to label drift typically involves human manual annotations and designed algorithms to assess annotation quality.
Modeling labeling quality control involves gathering a set of gold standard labels to assess and enhance the quality of annotation workers as well as other labels~\cite{jung2015modeling}.
To maintain continuous labeling quality~\cite{SIGIR2016}, it is important for multiple annotators to maintain consistent rules and criteria for the labeling procedure that can maintain inter-annotator variability. However, online detection of annotation bias is non-trivial and requires automatic algorithms for large-scale evaluation. 
Tracking the annotation workers on crowdsourcing platforms is essential to maintain quality and enable the modeling of labeling behaviors to facilitate automatic real-time monitoring~\cite{Goyal_McDonnell_Kutlu_Elsayed_Lease_2018}. Worker features are generated from their actions, including mouse clicks, scrolling, key presses, and focus changes. These features capture temporal behaviors by calculating event durations and time intervals between events. Certain attributes exhibit varying time scales influenced by human life experiences, biological status, and internal states. Preference modeling for recommendation systems is inherently shaped by personal thoughts and societal events. Time-aware recommendation systems utilize matrix factorization techniques to decompose and identify latent factors associated with temporal dynamics~\cite{zafari2019modelling}.
Human manual annotation of time series data involves continuous labeling within a streaming process, which is influenced by cognitive load and response delay. For instance, modeling physiological data~\cite{8657692} using aggregated labels from multiple annotators leads to improved performance.


\subsubsection{Human In The Loop}
\label{sssec:human_in_the_loop}

We now explore techniques relevant to human-in-the-loop model learning, which includes research on detecting human annotator performance and incorporating humans' next actions. In particular, active learning plays a significant role by providing data sampling and adaptive model learning approaches to guide the labeling of the next set of data in an iterative process. Initially, a model is trained with a small amount of data and selects a subset of data for the annotator to label, subsequently updating the model with the newly labeled data.


Developing effective active learning strategies ensures stable model learning while considering labeling, storage, and computational costs~\cite{9492291}. Key techniques in active learning include uncertainty-based, representation-based, and hybrid approaches~\cite{zhan2022comparative}. These approaches offer optimal strategies for efficient and cost-effective model learning.
Active learning strategies need to consider different data input forms. In the pool-based setting, a pool of unlabeled data is available, while the stream-based setting has limited choices of query samples due to the data being obtained in a stream. In terms of addressing label drift, the setting is similar to the stream-based setting. The pool-based approach can be seen as an adaptive strategy within the semi-supervised domain adaptation framework, involving humans in the process.
One prevalent issue in active learning approaches is the cold start problem, where the initial model has limited data to accurately select key unlabeled samples. Another human-engaged approach, prompt learning, also faces similar challenges when model updating is guided by inaccurate queries~\cite{yu2022cold}.
Recently, Hacohen {\em et al.} highlighted that low-budget and high-budget scenarios require different sample selection strategies~\cite{hacohen2022active}. In the low-budget case, actively selecting similar samples to the current dataset can aid model initialization. Conversely, the high-budget case involves the concept of hard-sample mining to enhance the model's discriminative power.

\subsection{Continual Learning}
\label{ssec:continual learning}

Catastrophic forgetting in data streams can result in significant failures in model prediction~\cite{korycki2021streaming}. To mitigate this issue, {\em continual learning}, also known as {\em life-long learning}, has been developed to update models without causing forgetting effects. Continual learning enables the consolidation of knowledge, allowing models to handle new information while retaining previously learned knowledge. One approach, Neural Network with Dynamically Evolved Capacity (NADINE), employs adjustable layers to perform partial updating of the network, preventing severe forgetting during updates~\cite{pratama2019automatic}. A survey on these techniques can be found in~\cite{ijcai2022p788}. Interestingly, in the context of long-term adaptation where accumulating vast amounts of new data is cumbersome, some forgetting can be beneficial~\cite{LIU2016322}. Consequently, the goal of these algorithms is to update new knowledge smoothly while minimizing the rapid loss of previously acquired knowledge.



The comprehensive exploration of continual learning frameworks in the context of concept drift is still relatively limited. Most existing research on continual concept drift primarily focuses on adapting to upcoming new instances using incremental learning techniques. However, the development of adapted models for new classes and tasks, in addition to handling domain differences over time, has received little attention~\cite{van2022three}.
Only a small number of recent studies have begun to unify continual learning with new classes in the concept drift framework, particularly for incrementally learning class-imbalanced new data~\cite{Korycki_2021_CVPR}. For example, Darem {\em et al.} utilize an incremental learning approach to integrate new malware detection capabilities without sacrificing the detection performance on existing instances~\cite{darem2021adaptive}. Similarly, Taufique {\em et al.} redesign domain shift datasets to simulate gradual changes inspired by the concept drift setting~\cite{Taufique_2022_CVPR}.
Continual learning serves as an intermediate research topic that bridges the fields of domain shift and concept drift, as researchers from both domains strive to advance knowledge in this area.


\begin{table*}[t]
\caption{
{\bf Techniques commonly used in each section.} 
Superscript number denotes the occurrence count of each technique in the literature. Referred to Fig.~\ref{fig:taxonomy} for the three-phase problem detection scheme and Table~\ref{tab:techniques} for the detailed organization of the techniques. Bold texts highlight techniques used in both fields of domain shift and concept drift.
\vspace{-2mm}
}
\label{tab:sections}
\addtolength{\tabcolsep}{-5pt}
\centerline{
\begin{tabular}{lll}
\toprule
\multicolumn{3}{c}{\textbf{Domain Shift}} \\
\midrule
Phase & Section                & Techniques           \\
\midrule
I & Out-of-Domain Detection    & Self-Supervised Learning$^5$~\cite{winkens2020contrastive,Mohseni_Pitale_Yadawa_Wang_2020}, \textbf{Distribution Learning}$^3$~\cite{pmlrOOD2022,9741317}, \textbf{Performance Monitoring}$^1$~\cite{Morteza_Li_2022} \\
\hline
II & \makecell[l]{Domain Generalization \\Domain Adaptation} & \makecell[l]{Data Manipulation$^3$~\cite{Ding_2022_CVPR}, Self-Supervised Learning$^5$~\cite{Xia_2021_ICCV,Kang_2019_CVPR}, \textbf{Regularization}$^4$~\cite{kim2021selfreg}, \\ Distribution Learning$^3$~\cite{ganin2016domain,Xu_Zhang_Ni_Li_Wang_Tian_Zhang_2020}, Feature Alignment$^2$~\cite{rozantsev2018beyond,sun2016return}, Sample Selection$^2$~\cite{chu2022denoised}, \\ Teacher-Student Learning$^3$~\cite{frikha2021towards,li2017large}, Knowledge Distillation$^1$~\cite{Ao_Li_Ling_2017}, \\\textbf{Partial Network Update}$^3$~\cite{niu2022domain}, \textbf{Network Architecture Design}$^1$~\cite{Yang_2023_WACV} }\\
\hline
III & Label Shift                      & \makecell[l]{Data Manipulation$^3$~\cite{9981099,yisource}, Self-Supervised Learning$^5$~\cite{9981099}, Regularization$^4$~\cite{azizzadenesheli2019regularized,yisource}, \\ Distribution Learning$^3$~\cite{yu2020label}, Teacher-Student Learning$^3$~\cite{han2021towards}, \\ \textbf{Sample Selection}$^2$~\cite{Yu_2021_CVPR}, \textbf{Loss Adjustment}$^2$~\cite{NEURIPS2021_5e6bd7a6} }              \\
\hline
III & Fast Adaptation                  & \makecell[l]{Data Manipulation$^3$~\cite{survey_fewshot2023}, Self-Supervised Learning$^5$~\cite{tian2021understanding}, Regularization$^4$~\cite{garrido2023on},\\ Meta Learning$^2$~\cite{Chi_2022_CVPR}, Teacher-Student Learning$^3$~\cite{hu2022switch}, \textbf{Partial Network Update}$^3$~\cite{9359505,pmlr_houlsby19a,li-liang-2021-prefix,lester2021power,hu2021lora} }              \\
\hline
III & \makecell[l]{Test-Time Training/\\Test-Time Adaptation}   & \makecell[l]{Self-Supervised Learning$^5$~\cite{sun19ttt,NEURIPS2021_b618c321}, Regularization$^4$~\cite{niu2022efficient, NEURIPS2021_b618c321}, Meta Learning$^2$~\cite{Chi_2021_CVPR,goyaltest}, \\Feature Alignment$^2$~\cite{su2022revisiting}, \textbf{Partial Network Update}$^3$~\cite{wang2022continual} \textbf{Loss Adjustment}$^2$~\cite{niu2022efficient}} \\ 
\midrule\midrule
\multicolumn{3}{c}{\textbf{Concept Drift}} \\
\midrule
Phase & Section   & Techniques \\
\midrule
I & Drift Detection    & \makecell[l]{\textbf{Distribution Learning}$^1$~\cite{zubarouglu2021data,liu2018accumulating,castellani2021task,li2022ddg}, Ensemble$^3$~\cite{9802893,9961859,castellani2021task,sidhu2015online}, \\ Meta Learning$^1$~\cite{anderson2019recurring,anderson2019recurring}, \textbf{Performance Monitoring}$^2$~\cite{BAYRAM2022108632,yu2018request}, Time Series Model$^1$~\cite{li2022ddg,9783029} } \\
\hline
II & Drift Adaptation   & \makecell[l]{\textbf{Regularization}$^2$~\cite{chalkidis2022improved}, Ensemble$^3$~\cite{9748034,9047166}, \textbf{Partial Network Update}$^2$~\cite{heusinger2022passive}, \textbf{Network Architecture Design}$^1$~\cite{9783029}}      \\
\hline
III & Label Drift        & \makecell[l]{Ensemble$^3$~\cite{li2020incremental}, Factorization$^1$~\cite{zafari2019modelling}, Performance Monitoring$^2$~\cite{jung2015modeling,Goyal_McDonnell_Kutlu_Elsayed_Lease_2018}, \\ \textbf{Sample Selection}$^1$~\cite{9492291,yu2022cold,hacohen2022active}, \textbf{Loss Adjustment}$^1$~\cite{chalkidis2022improved}}     \\ 
\hline
III & Continual Learning & \textbf{Partial Network Update}$^2$~\cite{pratama2019automatic} \\
\bottomrule
\end{tabular}
}
\end{table*}

\begin{table*}[t]
\caption{
{\bf Technical terms referred in the data change problems and their corresponding techniques.} The focused domain categorizes algorithms based on their emphasis on either pretraining (source domain) or finetuning (target domain). The feature and label columns represent the primary technical design spaces for learning, as depicted in Fig.~\ref{fig:space_learning}.
\vspace{-3mm}
}
\label{tab:techniques}
\centerline{
\addtolength{\tabcolsep}{-0pt}
\begin{tabular}{lcccl}
\toprule
\textbf{Technical Term}    & \textbf{Focused Domain} & \textbf{Feature} & \textbf{Label} & \textbf{Related Techniques}   \\ 
\midrule
Data Manipulation             & Source                     & \cmark                &                & Data Augmentation, Pseudo Labels, Batch Normalization \\
\hline
Self-Supervised Learning      & Source                     & \cmark                & \cmark              & Contrastive   Loss                                                  \\
\hline
Regularization                & Source                     &                  & \cmark              &  Stochastic Weights Averaging                                                                  \\
\hline
Distribution Learning         & Source                     & \cmark               &                & Generative   Model, Density Estimation, Clustering, Data Imputation \\
\hline
Ensemble                      & Source                     & \cmark         &  \cmark            & Mixture-of-Expert                                     \\
\hline
Meta-Learning                 & Source                     &                  & \cmark             & Auxiliary Learning                                                               \\
\hline
Time Series Model             & Source                     & \cmark             &                & Autoregression \\ 
\hline
Network Architecture Design & Source                     & \cmark             &                & Transformer \\ 
\hline
Feature Alignment             & Both                       & \cmark               &                & Cluster Matching                                                  \\
\hline
Factorization                 & Both                       & \cmark               &                & Disentangle Representation                                        \\
\hline
Adversarial Learning          & Both                       &              & \cmark           & Generative Adversarial Network,   Gradient Adversarial Learning                                \\
\hline
Performance Monitoring              & Both                       &                  & \cmark          & Statistical Test \\
\hline
Reinforcement Learning        & Target                     &                  & \cmark            &   Domain Adaptive Imitation Learning                                                    \\
\hline
Sample Selection              & Target                     & \cmark              &                & Active Learning                                                   \\
\hline
Teacher-Student Learning      & Target                     & \cmark              & \cmark              & Curriculum   Learning                                               \\
\hline
Knowledge Distillation        & Target                     & \cmark               & \cmark           &  Self-distillation, Subspace Learning                                                                  \\
\hline
Partial Network Update        & Target                     &                  & \cmark            & Prompt Learning, Prefix Learning, Adapter                         \\
\hline
Loss Adjustment & Target                     &                  & \cmark           & Uncertainty Estimation                                            \\
\hline
Continual Learning            & Target                     &                  & \cmark           & Life-Long Learning                                                \\
\bottomrule
\end{tabular}
}
\end{table*}

\section{Cross-Field Discussion of Data Change}
\label{sec:discussion}

We utilized the proposed {\em three-phase problem categorization scheme} to review research on domain shift and concept drift in $\S$\ref{sec:domain adaptation} and $\S$\ref{sec:concept drift}. We identified several overlapping techniques that have been recently developed in both fields, and we will discuss these in $\S$\ref{ssec:similar_techniques}, where we will frequenly make correspopndences to the respective sections. $\S$\ref{ssec:complentary_branches} presents distinct methods that address the original problems of each field, which may complement one another. To provide a comprehensive overview of the technology trends bridging these two fields within the context of data change, Table~\ref{tab:sections} presents a list of state-of-the-art algorithms.



\begin{figure*}[t]
\centerline{
\includegraphics[width=\linewidth]{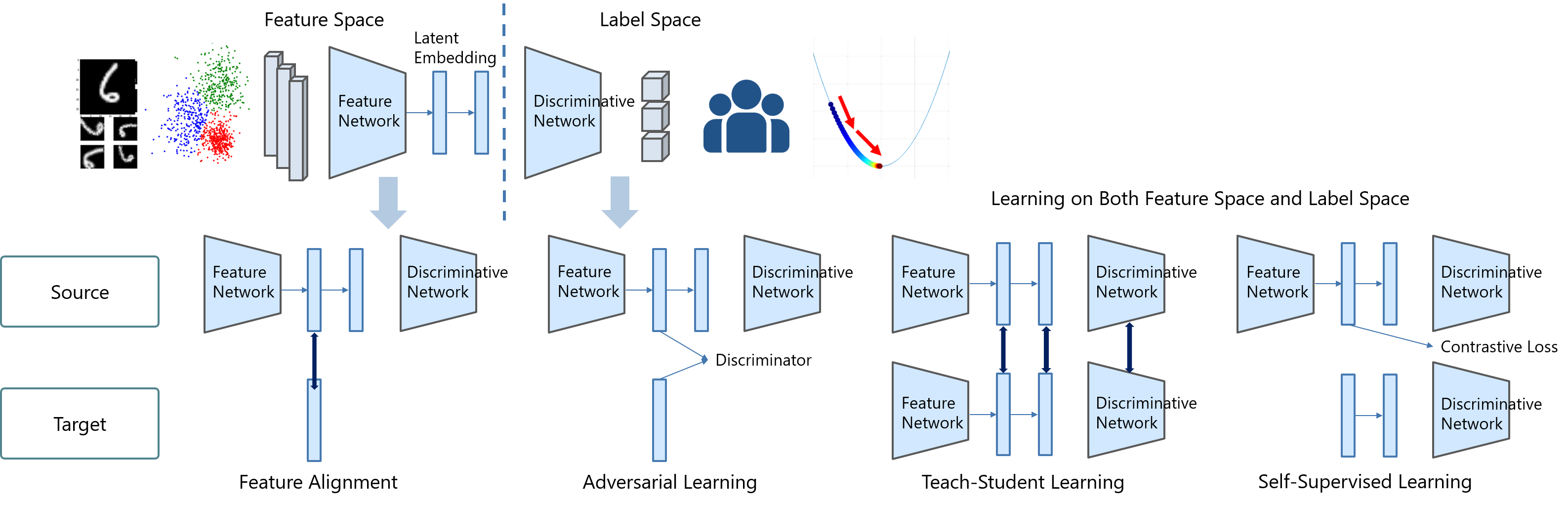}
\vspace{-3mm}
}
\caption{
{\bf Common learning schemes on the feature and label spaces.} 
(a) {\bf Feature alignment:} The encoder learns to map samples from different domains onto a shared latent space.
(b) {\bf Adversarial learning:} An additional discriminator is employed to distinguish between real and fake labels.
(c) {\bf Teacher-student learning:} Constraints are imposed on feature embeddings and classifiers.
(d) {\bf Self-supervised learning:} Pre-trained networks are generalized using contrastive loss to facilitate downstream tasks.
}
\label{fig:space_learning}
\end{figure*}

\subsection{Similar Underlying Structures and Techniques}
\label{ssec:similar_techniques}

The research topics presented in Fig.~\ref{fig:taxonomy} are categorized into three phases based on the shared structure of domain shift and concept drift problems. The first phase, {\em problem detection}, contains the OOD ($\S$\ref{ssec:OOD}) in-domain shift and drift detection ($\S$\ref{ssec:drift detection}) for concept drift. Once the data change is detected, the second phase, {\em problem-handling}, involves adaption techniques. Extensive efforts have been made in both domain shift ($\S$\ref{ssec:domain generalization}) and concept drift ($\S$\ref{ssec:drift_adaptation}) to adapt models for up-to-date usage. The third phase focuses on other factors, such as labeling issues recognized as label shift in the domain shift field ($\S$\ref{ssec:label_shift}) and label drift in the concept drift field ($\S$\ref{ssec:label drift}).
Time-related factors are emerging topics in domain shift, including Target Task Transfer (TTT) ($\S$\ref{ssec:ttt tta}) and Fast Adaptation ($\S$\ref{ssec:fast adaptation}), which incidentally align with concept drift for efficient model updating. Continual learning in concept drift ($\S$\ref{ssec:continual learning}) even leverages similar technical approaches as fast adaptation and TTT.

Table~\ref{tab:sections} provides an overview of cutting-edge techniques proposed in state-of-the-art studies. The listed methods offer valuable insights into addressing the challenges in domain shift and concept drift. Distribution learning and performance monitoring are common strategies employed during the detection phase. Regularization and partial network updating have proven to be effective approaches for adaptation in both domain shift and concept drift. Sample selection, particularly in the context of active learning, has been widely explored to mitigate label shift and label drift. Overall, both domain shift and concept drift offer a rich array of methods to address the negative effects of changes. The aforementioned techniques serve as general solutions applicable across different phases of these problems.

\subsection{Unique and Complementary Properties}
\label{ssec:complentary_branches}

The domains of domain shift and concept drift have historically emerged from separate communities, driven by distinct problem setups that naturally led to different intuitive directions. Domain shift arises when encountering a new batch of data, resulting in the development of frameworks centered around distribution-based learning. In Table~\ref{tab:sections}, self-supervised learning and teacher-student learning have been extensively developed, indicating a focus on distribution modeling.
Concept drift is characterized by the inherent streaming constraint, requiring sequential processes for detection, adaptation, and learning. Consequently, statistical-based performance monitoring, coupled with ensemble methods, offers a straightforward approach to robustly maintain an up-to-date model. Note that in Table~\ref{tab:sections}, distribution learning primarily pertains to drift detection rather than adaptation, as manipulating the entire distribution during adaptation can be costly given the streaming constraints.
These inherent differences in technical advancements reflect the original targeted problems within the respective fields, showcasing the unique challenges and solutions specific to the two fields.

Table~\ref{tab:techniques} and Fig.~\ref{fig:space_learning} illustrate the primary learning space of framework designs and major focused domains, providing guidance for selecting appropriate algorithms to address specific data change problems. Algorithms targeting the domain shift problem often consider constraints such as limited retraining resources and inaccessible source domain data. In contrast, approaches focusing on the source domain benefit from leveraging powerful pre-trained models trained on large-scale data to capture generic patterns. While the domain shift field typically emphasizes the source domain, recent studies have emerged that develop target domain-focused approaches built upon well-trained source domain models. Additionally, within the domain shift research, there is a growing consideration of temporal change as a factor influencing data distribution shift. On the other hand, the concept drift field leverages deep learning to expedite model performance. The potential to integrate these research advancements and reframe the problem is becoming increasingly evident.

\section{Industrial Perspective to Data Change}
\label{sec:industrial_perspective}

In dynamic environments of the industries, data undergo rapid changes aligned with market trends. $\S$~\ref{ssec:model_deployment} discusses the challenges associated with deploying models in such contexts. $\S$~\ref{ssec:applications} explores two real-world case studies.




\subsection{Model Deployment}
\label{ssec:model_deployment}

Deploying models for real-world industrial applications presents substantial challenges. Modern research efforts are addressing concerns regarding model size, data availability, and usability issues. These issues become more complicated as data undergoes change, creating new dimensions for model development.

\subsubsection{Efficiency Constraint}
\label{sssec:efficiency_constraint}

As the number of model parameters and the frequency of model training continue to grow, efficient management of resources becomes increasingly crucial. Solutions for domain shift and concept drift often necessitate the development of intricate frameworks to effectively handle rapid data changes over time. However, the trend of training large models has led to a significant increase in computational demands, raising concerns about sustainability. The research focus has shifted towards data-efficient learning, as highlighted in \cite{kuo2022green}. Cutting-edge advancements now underscore the significance of efficient frameworks, prioritizing factors such as low carbon footprints, compact model sizes, low computational complexity, and transparent logic. The objective is to achieve maximized model performance while incorporating reusable and traceable characteristics. Beyond enhancing model design, there is a growing emphasis on the human engagement throughout the model's usage process.


For the data change problem, a handful of research has laid the groundwork for efficient learning. Notably, techniques such as few-shot learning and foundation models discussed in $\S$\ref{ssec:fast adaptation} on the fast adaptation can be leveraged to reuse pre-trained models with little adaptation for reducing the required model size. However, the demand concerning timely adaptation, such as test-time adaptation ($\S$\ref{ssec:ttt tta}) and concept drift ($\S$\ref{sec:concept drift}), introduces a new dimension in resource consumption control. The cumulative consumption over time can become substantial. Hence, we emphasize that research fields focusing on temporal aspects should prioritize green learning as a primary direction in the future.


To enhance sustainability further, an important aspect is to harness human expertise, which directly aligns with the research field of human-in-the-loop ($\S$\ref{sssec:human_in_the_loop}). Current studies have explored efficient label acquisition, aiming to reduce human effort while improving model performance. Explainable AI emerges as a promising field that involves humans in the process, ensuring transparent results. By incorporating humans and machines in a loop, we envision the model being updated through human guidance and humans expanding their knowledge through systematic updates. Despite the progress made in addressing data change, there remains a limited number of studies that investigate interpretable reasons for changes and the resulting adaptations, as most evaluations still rely on prediction accuracy, leaving room for exploration in the interpretability.


The findings in our study reveals a growing convergence in the research fields of domain adaptation and concept drift. The use of pre-trained models and representation learning techniques in domain adaptation can complement the statistical tests and temporal modeling approaches employed in concept drift. The combination of transferable pre-training models and interpretable statistical analysis holds the potential to establish a new paradigm for trustworthy collaboration between humans and machines. Technical advancements in temporal efficiency can encourage fine-grained sustainability assessment. Contemporary research should adopt the accuracy, model size, and speed for the performance assessment regarding adaptation. A systematic metric to assess the level of human-machine collaboration in these solutions is still lacking.


\subsubsection{Model Robustness and Maintenance}
\label{sssec:infrastructure}


Model deployment presents a range of challenges for sustainability, specifically in terms of model robustness and iteration maintenance.
Building robust accuracy that can withstand diverse noises in features and labels is crucial for establishing trust in machine predictions. These noises can originate from natural variations or deliberate insertions.


The field of {\em learning from noisy labels} ($\S$\ref{ssec:label_shift}) is an example of robust knowledge extraction from noisy data in the source domain. Adversarial training algorithms are proposed to keep samples away from decision boundaries, preventing unforseeing changes in model predictions under adversarial perturbations~\cite{HUANG2020100270,ijcai2021p591}. 
The OOD challenges described in $\S$\ref{ssec:OOD} can be viewed as another kind of robustness volunability that might lead to unexpected results. 
Further research is needed to investigate the robustness of models against adversarial attacks and out-of-distribution (OOD) scenarios across various contexts. For example, prompting ChatGPT~\cite{wang2023robustness} with typos, distracted sentences, or abnormal prompts should not yield socially detrimental outcomes.

Implementing algorithms to address the challenges of data change relies heavily on the infrastructure, as insufficient deployment speed can lead to delays in detecting and adapting to these changes. 
Continuous Training (CT) is a concept established upon standard Machine Learning Operations (MLOps) to provide model retraining as a pipeline. When data change occurs, a series of retraining steps can directly update the model with new data for deployment.
Moreover, the development of algorithms is significantly influenced by public resources, such as open-sourced frameworks that ensure code consistency, as well as the collaborative efforts within communities.


\subsubsection{Usability}
\label{sssec:usability}

We organize the usability challenges related to data change into three main categories: {\em privacy}, {\em fairness}, and {\em decision-making schemes}. These factors significantly influence the user's willingness to adopt AI technologies. We divide the AI users into two groups: professional users and general users. Professional users leverage AI models to streamline their work tasks and prioritize decision-making schemes that enhance task value creation. General users are more inclined to enjoy AI services as long as privacy protection and fairness awareness designs are integrated.


\textbf{Privacy:} Maintaining privacy during data change is crucial due to the increased risks associated with reusing source domain data, adapting to target domain data, and the specific application scenarios. The level of privacy protection needed depends on the context. Most studies assume that public data from the source domain are readily available, while the target domain data are scarce and private.
Differential Privacy (DP)~\cite{jin2021differentially} and Federated Learning (FL)~\cite{peterson2019private} are two technical branches to tackle the data privacy and model privacy issues. 
Another research direction of {\em source-free domain adaptation} discussed in $\S$\ref{sssec:source_setting}, aims to enhance the privacy conditions by prohibiting access to the source domain data. This approach aligns with the principles of data-efficient learning discussed in $\S$\ref{sssec:efficiency_constraint}, which seeks to maximize performance while minimizing data usage. In terms of time considerations, FL has been explored for concept drift~\cite{9685083,casado2022concept}. Additionally, privacy leakage can be seen as a potential source of drift that requires detection and adaptation.



\textbf{Fairness:} The fairness issue is naturally tied to data change, as it involves biased data distributions. The bias can occur across various batches of data (domain shift) or data streams (concept drift). The main difficulty is on how best to define fairness among individuals or groups. 
A recent study~\cite{mukherjee2022domain} treats  fairness issues as problems of domain shift and concept drift. 
It is worth noting that data change introduces potential risks of unfair distributions, even when the source domain has been calibrated~\cite{wang2022robust}. 
Adversarial learning has emerged as a promising technique in addressing fairness in both domain adaptation and fairness representation learning, although the choice of objective function remains crucial in determining the final performance~\cite{madras2018learning}.
Therefore, identifying appropriate definitions and assessment metrics for fairness remains an ongoing research topic in the context of data change. It is important to consider fairness assessment during the timely detection and adaptation processes discussed in the concept drift section. While research has focused on addressing temporal dynamics for robust accuracy, the drift of fairness over time has been overlooked.




\textbf{Decision-Making Scheme:} The integration of automatic models has foundamentally transformed the decision making process.
When faced with rapidly changing data, how best to leverage data-driven AI models for decision making  becomes challenging. Studied reviewed in $\S$\ref{sssec:human_in_the_loop} have provided a basic process under active learning. However, existing human-machine collaboration schemes are limited to the enhancements of annotation strategies. In practice, guidelines for human feedbacks can greatly facilitate human engagement and effective acquisition of human expertise.


Our discussions regarding concept drift (specifically label drift in $\S$\ref{ssec:label drift}) imply that model prediction can come with  unawared bias that may corrupt the outcome. Therefore, the decision-making scheme involving humans in the loop should not be limited to a one-way feedback loop, but should instead encompass a mutually corrected iterative procedure.



Addressing the aforementioned topics is crucial for establishing trust for human users. While existing literature has made progress, the solutions and strategies to maintain usability during data change remains an ongoing endeavor.
It is evident that there is a need for additional evaluation metrics beyond accuracy-based measures, to assess the effectiveness of data change solutions. These issues will continue to be a focus of research as the fields of AI advance and evolve.



\begin{table}[t]
\caption{{\bf Common data change scenarios in real-world applications.}
\vspace{-3mm}
}
\label{tab:common_deployment_cases}
\centerline{
\addtolength{\tabcolsep}{2pt}
\begin{tabular}{lll}
\toprule
              & Domain Shift & Concept Drift \\
\midrule
Manufacturing & \begin{tabular}[c]{@{}l@{}}Product variability \\ Product line\\  Operating pipeline*\\ Machine configuration*\end{tabular} & \begin{tabular}[c]{@{}l@{}}Decision variability \\ Environment\end{tabular} \\
\midrule
Healthcare    & \begin{tabular}[c]{@{}l@{}}Disease category \\ Region, Race, Gene\\ Device, Equipment*\end{tabular} & \begin{tabular}[c]{@{}l@{}}Decision variability\\Population change    \\ Diagnosis rule*\end{tabular} \\
\midrule
Robotics      & Use case & Environment \\
\bottomrule
\end{tabular}
}
\vspace{2mm}
\footnotesize{* denotes that the cases rarely occur.}
\end{table}



\subsection{Real-world Applications}
\label{ssec:applications}

We categorize research works concerning real-world applications based on their task and algorithmic characteristics. We note that there is universal solution for all real-world scenarios. The broad range of applications include smart manufacturing~\cite{ZENISEK2021507,survey_drift2014}, smart healthcare~\cite{zhang2022transfer,kouw2018introduction}, smart automotive~\cite{survey_drift2014}, internet-of-things~\cite{lin2019concept,zhang2022transfer}, voice assistants~\cite{survey_drift2014}, and recommendation systems~\cite{8571222,8496795}. Data change occurs in different frequencies and degrades the learned model in various levels. 
We next focus discussion on smart manufacturing in $\S$\ref{sssec:smart manufacturing} and smart healthcare in $\S$\ref{sssec:medical_applications}. Table~\ref{tab:common_deployment_cases} lists common scenarios of data change in real-world scenarios.


\subsubsection{Smart Manufacturing}
\label{sssec:smart manufacturing}

Smart manufacturing tasks typically involve defect detection, process optimization, and inventory forecasting.
In Table~\ref{tab:common_deployment_cases}, common data change problems can pose significant challenges in manufacturing applications. For example, the introduction of new product lines and the variability within a product category result in domain shifts. Even different versions of the same product category exhibit domain shifts. Additionally, changes in the operation pipeline or machinery can lead to rare yet substantial domain shifts. The renewal of working policies or hardware infrastructures also contributes to dramatic shifts. Concept drift, on the other hand, is often associated with human factors such as tiredness or cognitive biases, leading to label drift. Environmental changes over time, such as background noise and lighting conditions, can be overlooked by humans.



It is important for smart manufacturing models to operate with high robustness\. While out-of-distribution events might be rare in the product line due to the unified pipeline control, it remains crucial to establish protective mechanisms to mitigate the potential impact of unknown data classes.

Industrial machines often possess superior computational power compared to consumer devices, allowing for the deployment of models without compromising the performance or accuracy. In such case, fairness and privacy issues are thus less influential as the users are not external individuals. Decision-making schemes would focus on incorporating automatic machine-assisted predictions with the human operating pipeline.




\subsubsection{Smart Healthcare}
\label{sssec:medical_applications}

In healthcare applications and clinical practice, stringent data quality control and meticulous model deployment requirements are important. Table~\ref{tab:common_deployment_cases} highlights various scenarios encountered in these contexts, such as diverse disease categories, regional disparities, and variations in equipment~\cite{thiagarajan2019understanding}. Most diseases exhibit subtypes characterized by distinct phenotypic attributes. Meanwhile, data collection across different regions introduces diverse subject groups with variations in race and genetics. Inconsistent expert opinions can lead to discrepant labels, resulting in label noise stemming from biased individual annotations~\cite{9674886}. These instances of domain shift contribute to the existence of datasets with varying properties.

Concept drift in healthcare encompasses not only variations in human decision-making but also changes in population dynamics over time~\cite{thiagarajan2019understanding}. An example of a relatively uncommon concept drift case is the update of diagnostic rules prompted by newly recognized consensus within the medical community.

We next consider a clinical scenario in a medical center to illustrate the data change impacts. The centralized services run on the server machines of the hospitcal servers typically come with high performance. On the other hand, the requirements regarding robustness, privacy, and fairness concerns become critical, as they can directly impact treatment feasibility. The effectiveness of AI models in clinical practice hinges on a well-designed decision-making scheme that incorporates medical specialists. To achieve successful integration into the clinical pipeline, interpretable AI and feedback mechanisms are preferred.

\section{Conclusion}
\label{sec:conclusion}


Ensuring the model's generalization capability is crucial for leveraging AI in real-world applications. The challenge of data change encompasses the study of domain shift and concept drift across various fields. This paper provides a systematic framework that guides readers in understanding these two fields through a unified three-phase problem categorization scheme. Despite the distinct research perspectives that have shaped the development of these fields, we have highlighted the presence of common underlying properties that can complement each other. By recognizing and leveraging these shared characteristics, valuable insights are obtained to better address the data change challenges.

Through an investigation of state-of-the-art studies, we have identified several trends across various technical fields. Notably, self-supervised learning has emerged as a key factor in enhancing models' capabilities to handle diverse data changes. Additionally, innovative designs in regularization approaches, including novel loss functions and network layer techniques, have proven effective. Temporal modeling techniques have gained significance, driven by the need for efficient solutions, leading to the emergence of new research branches.
Partial network updating strategies, such as prompt and adapter methods, have become valuable for adapting to changes and are closely associated with continual learning. More and more domain shift methods consider temporal change as a factor of data distribution shift. Regarding concept drift,  model performance keeps increasing as new deep learning approaches are developed. This progress opens up possibilities for integrating these developments and reframing the problem at hand.


This paper also addresses an important research question regarding the development of model deployment in applications. We emphasize the multifaceted challenges associated with model efficiency, robustness, and other usability concerns, including privacy, fairness, and decision-making schemes. By reviewing accessible datasets and benchmarking evaluations, this paper facilitates model deployment developments. We also select two industrial applications to carry out case studies and show the different needs among real-world applications. 



{\bf Future Directions.}
Our comprehensive cross-field review has identified important future directions for addressing the data change problem. Firstly, further research is needed to develop advanced modeling approaches that efficiently capture temporal dynamics. This involves exploring network architectures and adaptation strategies that minimize frequency and parameter quantity during network updating. Effective data change detection methods can indicate the optimal timing for adaptation, preventing unexpected model failures.
Secondly, achieving model generalization for adaptation to any data change is a major focus, leveraging self-supervised learning and continual learning techniques. The aim is to establish a large foundation model that reduces efforts for downstream tasks, aligning with efficient learning approaches.
Thirdly, incorporating humans in the loop is a critical but challenging next step. Understanding human behaviors in the context of label shift and drift problems is currently neglected, limiting the sustainability of real-world applications involving human interaction. Exploring a human-machine cooperation paradigm is essential for effectively addressing data change challenges.

Studies on domain shift have yielded remarkable results in modeling distribution, offering a foundation for constructing pre-trained models applicable to various tasks. In the realm of concept drift research, considerations of efficiency and explainability are key factors when designing methods for timely adaptation. This review paper uncovers both shared techniques and distinct characteristics between these two fields. We advocate for further research to advance reliable model learning, thereby accelerating the realization of trustworthy AI.

\bibliographystyle{IEEEtran}
\bibliography{main}

\begin{thebibliography}{100}
\providecommand{\url}[1]{#1}
\csname url@samestyle\endcsname
\providecommand{\newblock}{\relax}
\providecommand{\bibinfo}[2]{#2}
\providecommand{\BIBentrySTDinterwordspacing}{\spaceskip=0pt\relax}
\providecommand{\BIBentryALTinterwordstretchfactor}{4}
\providecommand{\BIBentryALTinterwordspacing}{\spaceskip=\fontdimen2\font plus
\BIBentryALTinterwordstretchfactor\fontdimen3\font minus
  \fontdimen4\font\relax}
\providecommand{\BIBforeignlanguage}[2]{{%
\expandafter\ifx\csname l@#1\endcsname\relax
\typeout{** WARNING: IEEEtran.bst: No hyphenation pattern has been}%
\typeout{** loaded for the language `#1'. Using the pattern for}%
\typeout{** the default language instead.}%
\else
\language=\csname l@#1\endcsname
\fi
#2}}
\providecommand{\BIBdecl}{\relax}
\BIBdecl

\bibitem{alvarez2021towards}
D.~Alvarez-Coello, D.~Wilms, A.~Bekan, and J.~M. G{\'o}mez, ``Towards a
  data-centric architecture in the automotive industry,'' \emph{Procedia
  Computer Science}, vol. 181, pp. 658--663, 2021.

\bibitem{8246999}
U.~Jayasinghe, A.~Otebolaku, T.-W. Um, and G.~M. Lee, ``Data centric trust
  evaluation and prediction framework for iot,'' in \emph{ITU K}, 2017, pp.
  1--7.

\bibitem{jarrahi2022principles}
M.~H. Jarrahi, A.~Memariani, and S.~Guha, ``The principles of data-centric ai
  (dcai),'' \emph{arXiv:2211.14611}, 2022.

\bibitem{zhang2022transfer}
L.~Zhang and X.~Gao, ``Transfer adaptation learning: A decade survey,''
  \emph{IEEE TNNLS}, 2022.

\bibitem{jiang2022transferability}
J.~Jiang, Y.~Shu, J.~Wang, and M.~Long, ``Transferability in deep learning: A
  survey,'' \emph{arXiv:2201.05867}, 2022.

\bibitem{tan2018survey}
C.~Tan, F.~Sun, T.~Kong, W.~Zhang, C.~Yang, and C.~Liu, ``A survey on deep
  transfer learning,'' in \emph{ICANN}.\hskip 1em plus 0.5em minus 0.4em\relax
  Springer, 2018, pp. 270--279.

\bibitem{kouw2018introduction}
W.~M. Kouw and M.~Loog, ``An introduction to domain adaptation and transfer
  learning,'' \emph{arXiv:1812.11806}, 2018.

\bibitem{hu2020no}
H.~Hu, M.~Kantardzic, and T.~S. Sethi, ``No free lunch theorem for concept
  drift detection in streaming data classification: A review,'' \emph{WIREs
  DMKD}, vol.~10, no.~2, p. e1327, 2020.

\bibitem{wares2019data}
S.~Wares, J.~Isaacs, and E.~Elyan, ``Data stream mining: methods and challenges
  for handling concept drift,'' \emph{SN Applied Sciences}, vol.~1, pp. 1--19,
  2019.

\bibitem{8571222}
A.~S. Iwashita and J.~P. Papa, ``An overview on concept drift learning,''
  \emph{IEEE Access}, vol.~7, pp. 1532--1547, 2019.

\bibitem{8496795}
J.~Lu, A.~Liu, F.~Dong, F.~Gu, J.~Gama, and G.~Zhang, ``Learning under concept
  drift: A review,'' \emph{IEEE Transactions on Knowledge and Data
  Engineering}, vol.~31, no.~12, pp. 2346--2363, 2019.

\bibitem{9544693}
P.~A. Nayak, P.~Sriganesh, K.~Rakshitha, M.~Manoj~Kumar, B.~S. Prashanth, and
  H.~R. Sneha, ``Literature review on phenomenon of concept drift and its
  handling approaches,'' in \emph{ASIANCON}, 2021, pp. 1--7.

\bibitem{hashmani2020concept}
M.~A. Hashmani, S.~M. Jameel, M.~Rehman, and A.~Inoue, ``Concept drift
  evolution in machine learning approaches: a systematic literature review,''
  \emph{International Journal on Smart Sensing and Intelligent Systems},
  vol.~13, no.~1, pp. 1--16, 2020.

\bibitem{farahani2021brief}
A.~Farahani, S.~Voghoei, K.~Rasheed, and H.~R. Arabnia, ``A brief review of
  domain adaptation,'' \emph{ICDATA and IKE}, pp. 877--894, 2021.

\bibitem{ganin2016domain}
Y.~Ganin, E.~Ustinova, H.~Ajakan, P.~Germain \emph{et~al.},
  ``Domain-adversarial training of neural networks,'' \emph{The journal of
  machine learning research}, vol.~17, no.~1, pp. 2096--2030, 2016.

\bibitem{Xu_Zhang_Ni_Li_Wang_Tian_Zhang_2020}
M.~Xu, J.~Zhang, B.~Ni, T.~Li, C.~Wang, Q.~Tian, and W.~Zhang, ``Adversarial
  domain adaptation with domain mixup,'' \emph{AAAI Conference Proceedings},
  vol.~34, no.~04, pp. 6502--6509, 2020.

\bibitem{yu2020label}
X.~Yu, T.~Liu, M.~Gong, K.~Zhang, K.~Batmanghelich, and D.~Tao, ``Label-noise
  robust domain adaptation,'' in \emph{ICML}.\hskip 1em plus 0.5em minus
  0.4em\relax PMLR, 2020, pp. 10\,913--10\,924.

\bibitem{lesort2021understanding}
T.~Lesort, M.~Caccia, and I.~Rish, ``Understanding continual learning settings
  with data distribution drift analysis,'' \emph{arXiv:2104.01678}, 2021.

\bibitem{9124702}
P.~Ceravolo, G.~M. Tavares, S.~B. Junior, and E.~Damiani, ``Evaluation goals
  for online process mining: A concept drift perspective,'' \emph{IEEE
  Transactions on Services Computing}, vol.~15, no.~4, pp. 2473--2489, 2022.

\bibitem{8861136}
W.~M. Kouw and M.~Loog, ``A review of domain adaptation without target
  labels,'' \emph{IEEE Transactions on Pattern Analysis and Machine
  Intelligence}, vol.~43, no.~3, pp. 766--785, 2021.

\bibitem{SIP-2022-0019}
X.~Liu, C.~Yoo, F.~Xing, H.~Oh, G.~E. Fakhri, J.-W. Kang, and J.~Woo, ``Deep
  unsupervised domain adaptation: A review of recent advances and
  perspectives,'' \emph{APSIPA Transactions on Signal and Information
  Processing}, vol.~11, no.~1, pp.~--, 2022.

\bibitem{9238468}
S.~Zhao, X.~Yue, S.~Zhang, B.~Li, H.~Zhao \emph{et~al.}, ``A review of
  single-source deep unsupervised visual domain adaptation,'' \emph{IEEE
  Transactions on Neural Networks and Learning Systems}, vol.~33, no.~2, pp.
  473--493, 2022.

\bibitem{9289802}
N.~L.~A. Ghani, I.~A. Aziz, and M.~Mehat, ``Concept drift detection on
  unlabeled data streams: A systematic literature review,'' in \emph{ICBDA},
  2020, pp. 61--65.

\bibitem{Fahy2022}
C.~Fahy, S.~Yang, and M.~Gongora, ``Scarcity of labels in non-stationary data
  streams: A survey,'' \emph{ACM Comput. Surv.}, vol.~55, no.~2, 2022.

\bibitem{AGRAHARI20229523}
S.~Agrahari and A.~K. Singh, ``Concept drift detection in data stream mining :
  A literature review,'' \emph{Journal of King Saud University - Computer and
  Information Sciences}, vol.~34, no. 10, Part B, pp. 9523--9540, 2022.

\bibitem{yu2023comprehensive}
Z.~Yu, J.~Li, Z.~Du, L.~Zhu, and H.~T. Shen, ``A comprehensive survey on
  source-free domain adaptation,'' \emph{arXiv:2302.11803}, 2023.

\bibitem{SUAREZCETRULO2023118934}
A.~L. Suárez-Cetrulo, D.~Quintana, and A.~Cervantes, ``A survey on machine
  learning for recurring concept drifting data streams,'' \emph{Expert Systems
  with Applications}, vol. 213, p. 118934, 2023.

\bibitem{jameel2020critical}
S.~M. Jameel, M.~A. Hashmani, H.~Alhussain, M.~Rehman, and A.~Budiman, ``A
  critical review on adverse effects of concept drift over machine learning
  classification models,'' \emph{IJACSA}, vol.~11, no.~1, 2020.

\bibitem{zhou2022domain}
K.~Zhou, Z.~Liu, Y.~Qiao, T.~Xiang, and C.~C. Loy, ``Domain generalization: A
  survey,'' \emph{IEEE PAMI}, no.~01, pp. 1--20, 2022.

\bibitem{9762269}
M.~Lima, M.~Neto, T.~S. Filho, and R.~A. de~A.~Fagundes, ``Learning under
  concept drift for regression—a systematic literature review,'' \emph{IEEE
  Access}, vol.~10, pp. 45\,410--45\,429, 2022.

\bibitem{SurveyMin2022}
M.~Fan, Z.~Cai, T.~Zhang, and B.~Wang, ``A survey of deep domain adaptation
  based on label set classification,'' \emph{Multimedia Tools Appl.}, vol.~81,
  no.~27, p. 39545–39576, 2022.

\bibitem{8246564}
S.~Wang, L.~L. Minku, and X.~Yao, ``A systematic study of online class
  imbalance learning with concept drift,'' \emph{IEEE TNNLS}, vol.~29, no.~10,
  pp. 4802--4821, 2018.

\bibitem{saunders2022domain}
D.~Saunders, ``Domain adaptation and multi-domain adaptation for neural machine
  translation: A survey,'' \emph{Journal of Artificial Intelligence Research},
  vol.~75, pp. 351--424, 2022.

\bibitem{guo2022domain}
X.~Guo and H.~Yu, ``On the domain adaptation and generalization of pretrained
  language models: A survey,'' \emph{arXiv:2211.03154}, 2022.

\bibitem{chan2023state}
J.~Y.-L. Chan, K.~T. Bea, S.~M.~H. Leow, S.~W. Phoong, and W.~K. Cheng, ``State
  of the art: a review of sentiment analysis based on sequential transfer
  learning,'' \emph{Artificial Intelligence Review}, vol.~56, no.~1, pp.
  749--780, 2023.

\bibitem{ijcai2022p788}
L.~Yuan, H.~Li, B.~Xia, C.~Gao, M.~Liu, W.~Yuan, and X.~You, ``Recent advances
  in concept drift adaptation methods for deep learning,'' in \emph{IJCAI},
  2022, pp. 5654--5661.

\bibitem{yadav2022survey}
H.~Yadav and S.~Sitaram, ``A survey of multilingual models for automatic speech
  recognition,'' in \emph{LREC}, 2022, pp. 5071--5079.

\bibitem{9296327}
P.~Bell, J.~Fainberg, O.~Klejch, J.~Li, S.~Renals, and P.~Swietojanski,
  ``Adaptation algorithms for neural network-based speech recognition: An
  overview,'' \emph{IEEE Open Journal of Signal Processing}, vol.~2, pp.
  33--66, 2021.

\bibitem{BAYRAM2022108632}
F.~Bayram, B.~S. Ahmed, and A.~Kassler, ``From concept drift to model
  degradation: An overview on performance-aware drift detectors,''
  \emph{Knowledge-Based Systems}, vol. 245, p. 108632, 2022.

\bibitem{zhu2020transfer}
Z.~Zhu, K.~Lin, and J.~Zhou, ``Transfer learning in deep reinforcement
  learning: A survey,'' \emph{arXiv:2009.07888}, 2020.

\bibitem{mahmood2021concept}
T.~Mahmood and T.~Fatima, ``Concept drift in streaming data: A systematic
  literature review,'' \emph{KIET Journal of Computing and Information
  Sciences}, vol.~4, no.~1, pp. 17--17, 2021.

\bibitem{arjovsky2019invariant}
M.~Arjovsky, L.~Bottou, I.~Gulrajani, and D.~Lopez-Paz, ``Invariant risk
  minimization,'' \emph{arXiv:1907.02893}, 2019.

\bibitem{ghifary2015domain}
M.~Ghifary, W.~B. Kleijn, M.~Zhang, and D.~Balduzzi, ``Domain generalization
  for object recognition with multi-task autoencoders,'' in \emph{ICCV}, 2015,
  pp. 2551--2559.

\bibitem{hendrycks2019benchmarking}
D.~Hendrycks and T.~Dietterich, ``Benchmarking neural network robustness to
  common corruptions and perturbations,'' in \emph{ICLR}, 2019.

\bibitem{sun2022benchmarking}
J.~Sun, Q.~Zhang, B.~Kailkhura, Z.~Yu, C.~Xiao, and Z.~M. Mao, ``Benchmarking
  robustness of 3d point cloud recognition against common corruptions,''
  \emph{arXiv:2201.12296}, 2022.

\bibitem{visda2017}
X.~Peng, B.~Usman, N.~Kaushik, J.~Hoffman, D.~Wang, and K.~Saenko, ``Visda: The
  visual domain adaptation challenge,'' \emph{arXiv:1710.06924}, 2017.

\bibitem{pmlr19b_animaln}
H.~Song, M.~Kim, and J.-G. Lee, ``{SELFIE}: Refurbishing unclean samples for
  robust deep learning,'' in \emph{ICML}, vol.~97.\hskip 1em plus 0.5em minus
  0.4em\relax PMLR, 2019, pp. 5907--5915.

\bibitem{wei2022learning}
J.~Wei, Z.~Zhu, H.~Cheng, T.~Liu, G.~Niu, and Y.~Liu, ``Learning with noisy
  labels revisited: A study using real-world human annotations,'' in
  \emph{ICLR}, 2022.

\bibitem{xiao2015learning}
T.~Xiao, T.~Xia, Y.~Yang, C.~Huang, and X.~Wang, ``Learning from massive noisy
  labeled data for image classification,'' in \emph{CVPR}, 2015, pp.
  2691--2699.

\bibitem{li2017webvision}
W.~Li, L.~Wang, W.~Li, E.~Agustsson, and L.~Van~Gool, ``Webvision database:
  Visual learning and understanding from web data,'' \emph{arXiv:1708.02862},
  2017.

\bibitem{li2017deeper}
D.~Li, Y.~Yang, Y.-Z. Song, and T.~M. Hospedales, ``Deeper, broader and artier
  domain generalization,'' in \emph{ICCV}, 2017, pp. 5542--5550.

\bibitem{venkateswara2017deep}
H.~Venkateswara, J.~Eusebio, S.~Chakraborty, and S.~Panchanathan, ``Deep
  hashing network for unsupervised domain adaptation,'' in \emph{CVPR}, 2017,
  pp. 5018--5027.

\bibitem{zhao2019multi}
S.~Zhao, B.~Li, X.~Yue, Y.~Gu, P.~Xu, R.~Hu, H.~Chai, and K.~Keutzer,
  ``Multi-source domain adaptation for semantic segmentation,'' \emph{NeurIPS},
  vol.~32, 2019.

\bibitem{hendrycks2021many}
D.~Hendrycks, S.~Basart, N.~Mu, S.~Kadavath, F.~Wang, E.~Dorundo, R.~Desai,
  T.~Zhu, S.~Parajuli, M.~Guo \emph{et~al.}, ``The many faces of robustness: A
  critical analysis of out-of-distribution generalization,'' in \emph{ICCV},
  2021, pp. 8340--8349.

\bibitem{fang2013unbiased}
C.~Fang, Y.~Xu, and D.~N. Rockmore, ``Unbiased metric learning: On the
  utilization of multiple datasets and web images for softening bias,'' in
  \emph{ICCV}, 2013, pp. 1657--1664.

\bibitem{koh2021wilds}
P.~W. Koh, S.~Sagawa, H.~Marklund, S.~M. Xie, M.~Zhang \emph{et~al.}, ``Wilds:
  A benchmark of in-the-wild distribution shifts,'' in \emph{ICML}.\hskip 1em
  plus 0.5em minus 0.4em\relax PMLR, 2021, pp. 5637--5664.

\bibitem{he2021towards}
Y.~He, Z.~Shen, and P.~Cui, ``Towards non-iid image classification: A dataset
  and baselines,'' \emph{Pattern Recognition}, vol. 110, p. 107383, 2021.

\bibitem{nǎdejde2022cocoa}
M.~Nǎdejde, A.~Currey, B.~Hsu, X.~Niu, M.~Federico, and G.~Dinu, ``Cocoa-mt: A
  dataset and benchmark for contrastive controlled mt with application to
  formality,'' in \emph{NAACL}, 2022, pp. 616--632.

\bibitem{9052492}
Y.~Zheng, G.~Chen, and M.~Huang, ``Out-of-domain detection for natural language
  understanding in dialog systems,'' \emph{IEEE/ACM Transactions on Audio,
  Speech, and Language Processing}, vol.~28, pp. 1198--1209, 2020.

\bibitem{coucke2018snips}
A.~Coucke, A.~Saade, A.~Ball, T.~Bluche, A.~Caulier \emph{et~al.}, ``Snips
  voice platform: an embedded spoken language understanding system for
  private-by-design voice interfaces,'' \emph{arXiv:1805.10190}, 2018.

\bibitem{larson-etal-2019-evaluation}
S.~Larson, A.~Mahendran, J.~J. Peper, C.~Clarke, A.~Lee \emph{et~al.}, ``An
  evaluation dataset for intent classification and out-of-scope prediction,''
  in \emph{EMNLP-IJCNLP}, 2019.

\bibitem{zang-etal-2020-multiwoz}
X.~Zang, A.~Rastogi, S.~Sunkara, R.~Gupta, J.~Zhang, and J.~Chen,
  ``{M}ulti{WOZ} 2.2 : A dialogue dataset with additional annotation
  corrections and state tracking baselines,'' in \emph{NLP4ConvAI}, 2020, pp.
  109--117.

\bibitem{walker2006ace}
C.~Walker, S.~Strassel, J.~Medero, and K.~Maeda, ``Ace 2005 multilingual
  training corpus,'' \emph{Linguistic Data Consortium, Philadelphia}, vol.~57,
  p.~45, 2006.

\bibitem{blitzer-etal-2007-biographies}
J.~Blitzer, M.~Dredze, and F.~Pereira, ``Biographies, {B}ollywood, boom-boxes
  and blenders: Domain adaptation for sentiment classification,'' in
  \emph{ACL}, 2007, pp. 440--447.

\bibitem{9053074}
A.~Mathur, F.~Kawsar, N.~Berthouze, and N.~D. Lane, ``Libri-adapt: a new speech
  dataset for unsupervised domain adaptation,'' in \emph{ICASSP}, 2020, pp.
  7439--7443.

\bibitem{commonvoice:2020}
R.~Ardila, M.~Branson, K.~Davis, M.~Henretty, M.~Kohler \emph{et~al.}, ``Common
  voice: A massively-multilingual speech corpus,'' in \emph{LREC 2020}, 2020,
  pp. 4211--4215.

\bibitem{karouzos-etal-2021-udalm}
C.~Karouzos, G.~Paraskevopoulos, and A.~Potamianos, ``{UDALM}: Unsupervised
  domain adaptation through language modeling,'' in \emph{NAACL}, 2021, pp.
  2579--2590.

\bibitem{chu2022denoised}
T.~Chu, Y.~Liu, J.~Deng, W.~Li, and L.~Duan, ``Denoised maximum classifier
  discrepancy for source-free unsupervised domain adaptation,'' in \emph{AAAI},
  vol.~36, no.~1, 2022, pp. 472--480.

\bibitem{Ao_Li_Ling_2017}
S.~Ao, X.~Li, and C.~Ling, ``Fast generalized distillation for semi-supervised
  domain adaptation,'' in \emph{AAAI}, vol.~31, no.~1, 2017.

\bibitem{8273627}
C.-M. Chang, B.-H. Su, S.-C. Lin, J.-L. Li, and C.-C. Lee, ``A bootstrapped
  multi-view weighted kernel fusion framework for cross-corpus integration of
  multimodal emotion recognition,'' in \emph{ACII}, 2017, pp. 377--382.

\bibitem{lu2018learning}
C.-C. Lu, J.-L. Li, and C.-C. Lee, ``Learning an arousal-valence speech
  front-end network using media data in-the-wild for emotion recognition,'' in
  \emph{AVEC}, 2018, pp. 99--105.

\bibitem{5975223}
R.~Elwell and R.~Polikar, ``Incremental learning of concept drift in
  nonstationary environments,'' \emph{IEEE Transactions on Neural Networks},
  vol.~22, no.~10, pp. 1517--1531, 2011.

\bibitem{zliobaite2013good}
I.~Zliobaite, ``How good is the electricity benchmark for evaluating concept
  drift adaptation,'' \emph{arXiv:1301.3524}, 2013.

\bibitem{blackard1999comparative}
J.~A. Blackard and D.~J. Dean, ``Comparative accuracies of artificial neural
  networks and discriminant analysis in predicting forest cover types from
  cartographic variables,'' \emph{Computers and electronics in agriculture},
  vol.~24, no.~3, pp. 131--151, 1999.

\bibitem{7280610}
V.~Losing, B.~Hammer, and H.~Wersing, ``Interactive online learning for
  obstacle classification on a mobile robot,'' in \emph{IJCNN}, 2015, pp. 1--8.

\bibitem{7837853}
------, ``Knn classifier with self adjusting memory for heterogeneous concept
  drift,'' in \emph{ICDM}, 2016, pp. 291--300.

\bibitem{vzliobaite2011combining}
I.~{\v{Z}}liobait{\.e}, ``Combining similarity in time and space for training
  set formation under concept drift,'' \emph{Intelligent Data Analysis},
  vol.~15, no.~4, pp. 589--611, 2011.

\bibitem{souza2020challenges}
V.~M. Souza, D.~M. dos Reis, A.~G. Maletzke, and G.~E. Batista, ``Challenges in
  benchmarking stream learning algorithms with real-world data,'' \emph{Data
  Mining and Knowledge Discovery}, vol.~34, pp. 1805--1858, 2020.

\bibitem{yang2021oodsurvey}
J.~Yang, K.~Zhou, Y.~Li, and Z.~Liu, ``Generalized out-of-distribution
  detection: A survey,'' \emph{arXiv:2110.11334}, 2021.

\bibitem{Morteza_Li_2022}
P.~Morteza and Y.~Li, ``Provable guarantees for understanding
  out-of-distribution detection,'' in \emph{AAAI}, vol.~36, no.~7, 2022, pp.
  7831--7840.

\bibitem{NEURIPS2021_01894d6f}
Y.~Sun, C.~Guo, and Y.~Li, ``{ReAct:} out-of-distribution detection with
  rectified activations,'' in \emph{NeurIPS}, vol.~34, 2021, pp. 144--157.

\bibitem{pmlrOOD2022}
Y.~Sun, Y.~Ming, X.~Zhu, and Y.~Li, ``Out-of-distribution detection with deep
  nearest neighbors,'' in \emph{ICML}, vol. 162.\hskip 1em plus 0.5em minus
  0.4em\relax PMLR, 2022, pp. 20\,827--20\,840.

\bibitem{winkens2020contrastive}
J.~Winkens, R.~Bunel, A.~G. Roy, R.~Stanforth, V.~Natarajan \emph{et~al.},
  ``Contrastive training for improved out-of-distribution detection,''
  \emph{arXiv:2007.05566}, 2020.

\bibitem{Mohseni_Pitale_Yadawa_Wang_2020}
S.~Mohseni, M.~Pitale, J.~Yadawa, and Z.~Wang, ``Self-supervised learning for
  generalizable out-of-distribution detection,'' in \emph{AAAI}, vol.~34,
  no.~04, 2020, pp. 5216--5223.

\bibitem{du2022siren}
X.~Du, G.~Gozum, Y.~Ming, and Y.~Li, ``{SIREN}: Shaping representations for
  detecting out-of-distribution objects,'' in \emph{NeurIPS}, 2022.

\bibitem{9741317}
D.~Jin, S.~Gao, S.~Kim, Y.~Liu, and D.~Hakkani-Tür, ``Towards textual
  out-of-domain detection without in-domain labels,'' \emph{IEEE/ACM
  Transactions on Audio, Speech, and Language Processing}, vol.~30, pp.
  1386--1395, 2022.

\bibitem{yangopenood}
J.~Zhang, J.~Yang, P.~Wang, H.~Wang, Y.~Lin \emph{et~al.}, ``{OpenOOD} v1.5:
  Enhanced benchmark for out-of-distribution detection,'' in \emph{arXiv
  preprint arXiv:2306.09301}, 2023.

\bibitem{Kirchheim_2022_CVPR}
K.~Kirchheim, M.~Filax, and F.~Ortmeier, ``{PyTorch-OOD}: A library for
  out-of-distribution detection based on {PyTorch},'' in \emph{CVPR Workshops},
  2022, pp. 4351--4360.

\bibitem{wang2022generalizing}
J.~Wang, C.~Lan, C.~Liu, Y.~Ouyang, T.~Qin \emph{et~al.}, ``Generalizing to
  unseen domains: A survey on domain generalization,'' \emph{IEEE Transactions
  on Knowledge and Data Engineering}, 2022.

\bibitem{Qiao_2020_CVPR}
F.~Qiao, L.~Zhao, and X.~Peng, ``Learning to learn single domain
  generalization,'' in \emph{CVPR}, 2020.

\bibitem{Wang_2022_CVPR}
F.~Wang, Z.~Han, Y.~Gong, and Y.~Yin, ``Exploring domain-invariant parameters
  for source free domain adaptation,'' in \emph{CVPR}, 2022, pp. 7151--7160.

\bibitem{9200758}
Y.~Kim, D.~Cho, and S.~Hong, ``Towards privacy-preserving domain adaptation,''
  \emph{IEEE Signal Processing Letters}, vol.~27, pp. 1675--1679, 2020.

\bibitem{Ding_2022_CVPR}
N.~Ding, Y.~Xu, Y.~Tang, C.~Xu, Y.~Wang, and D.~Tao, ``Source-free domain
  adaptation via distribution estimation,'' in \emph{CVPR}, 2022, pp.
  7212--7222.

\bibitem{Xia_2021_ICCV}
H.~Xia, H.~Zhao, and Z.~Ding, ``Adaptive adversarial network for source-free
  domain adaptation,'' in \emph{ICCV}, 2021, pp. 9010--9019.

\bibitem{frikha2021towards}
A.~Frikha, H.~Chen, D.~Krompa{\ss}, T.~Runkler, and V.~Tresp, ``Towards
  data-free domain generalization,'' in \emph{NeurIPS Workshop}, 2021.

\bibitem{niu2022domain}
H.~Niu, H.~Li, F.~Zhao, and B.~Li, ``Domain-unified prompt representations for
  source-free domain generalization,'' \emph{arXiv:2209.14926}, 2022.

\bibitem{li2017large}
J.~Li, M.~L. Seltzer, X.~Wang, R.~Zhao, and Y.~Gong, ``Large-scale domain
  adaptation via teacher-student learning,'' \emph{Proc. Interspeech}, pp.
  2386--2390, 2017.

\bibitem{kim2021selfreg}
D.~Kim, Y.~Yoo, S.~Park, J.~Kim, and J.~Lee, ``{SelfReg}: Self-supervised
  contrastive regularization for domain generalization,'' in \emph{ICCV}, 2021,
  pp. 9619--9628.

\bibitem{Tang_Jia_2020}
H.~Tang and K.~Jia, ``Discriminative adversarial domain adaptation,'' in
  \emph{AAAI}, vol.~34, 2020, pp. 5940--5947.

\bibitem{Yang_2023_WACV}
J.~Yang, J.~Liu, N.~Xu, and J.~Huang, ``Tvt: Transferable vision transformer
  for unsupervised domain adaptation,'' in \emph{WACV}, 2023, pp. 520--530.

\bibitem{gideon2019improving}
J.~Gideon, M.~G. McInnis, and E.~M. Provost, ``Improving cross-corpus speech
  emotion recognition with adversarial discriminative domain generalization
  ({ADDoG}),'' \emph{IEEE Transactions on Affective Computing}, vol.~12, no.~4,
  pp. 1055--1068, 2019.

\bibitem{rozantsev2018beyond}
A.~Rozantsev, M.~Salzmann, and P.~Fua, ``Beyond sharing weights for deep domain
  adaptation,'' \emph{IEEE PAMI}, vol.~41, no.~4, pp. 801--814, 2018.

\bibitem{sun2016return}
B.~Sun, J.~Feng, and K.~Saenko, ``Return of frustratingly easy domain
  adaptation,'' in \emph{AAAI}, vol.~30, no.~1, 2016.

\bibitem{Kang_2019_CVPR}
G.~Kang, L.~Jiang, Y.~Yang, and A.~G. Hauptmann, ``Contrastive adaptation
  network for unsupervised domain adaptation,'' in \emph{CVPR}, 2019.

\bibitem{shen2018wasserstein}
J.~Shen, Y.~Qu, W.~Zhang, and Y.~Yu, ``Wasserstein distance guided
  representation learning for domain adaptation,'' in \emph{AAAI}, vol.~32,
  no.~1, 2018.

\bibitem{li2016revisiting}
Y.~Li, N.~Wang, J.~Shi, J.~Liu, and X.~Hou, ``Revisiting batch normalization
  for practical domain adaptation,'' \emph{arXiv:1603.04779}, 2016.

\bibitem{lipton2018detecting}
Z.~Lipton, Y.-X. Wang, and A.~Smola, ``Detecting and correcting for label shift
  with black box predictors,'' in \emph{ICML}.\hskip 1em plus 0.5em minus
  0.4em\relax PMLR, 2018, pp. 3122--3130.

\bibitem{azizzadenesheli2019regularized}
K.~Azizzadenesheli, A.~Liu, F.~Yang, and A.~Anandkumar, ``Regularized learning
  for domain adaptation under label shifts,'' in \emph{ICLR}, 2019.

\bibitem{chen2022unsupervised}
L.~Chen, M.~Zaharia, and J.~Y. Zou, ``Is unsupervised performance estimation
  impossible when both covariates and labels shift?'' in \emph{NeurIPS
  Workshop}, 2022.

\bibitem{NEURIPS2021_5e6bd7a6}
R.~Wu, C.~Guo, Y.~Su, and K.~Q. Weinberger, ``Online adaptation to label
  distribution shift,'' in \emph{NeurIPS}, vol.~34, 2021, pp. 11\,340--11\,351.

\bibitem{song2022learning}
H.~Song, M.~Kim, D.~Park, Y.~Shin, and J.-G. Lee, ``Learning from noisy labels
  with deep neural networks: A survey,'' \emph{IEEE TNNLS}, 2022.

\bibitem{kumar22_interspeech}
A.~Kumar, P.~K. Sharma, A.~Illa, S.~Venkatapathy \emph{et~al.}, ``Learning
  under label noise for robust spoken language understanding systems,'' in
  \emph{Proc. Interspeech 2022}, 2022, pp. 3463--3467.

\bibitem{han2021towards}
Z.~Han, X.-J. Gui, C.~Cui, and Y.~Yin, ``Towards accurate and robust domain
  adaptation under noisy environments,'' in \emph{IJCAI}, 2021, pp. 2269--2276.

\bibitem{Yu_2021_CVPR}
Q.~Yu, A.~Hashimoto, and Y.~Ushiku, ``Divergence optimization for noisy
  universal domain adaptation,'' in \emph{CVPR}, 2021, pp. 2515--2524.

\bibitem{9981099}
W.~Chen, L.~Lin, S.~Yang, D.~Xie, S.~Pu, and Y.~Zhuang, ``Self-supervised noisy
  label learning for source-free unsupervised domain adaptation,'' in
  \emph{IROS}, 2022, pp. 10\,185--10\,192.

\bibitem{yisource}
L.~Yi, G.~Xu, P.~Xu, J.~Li, R.~Pu, C.~Ling, I.~McLeod, and B.~Wang, ``When
  source-free domain adaptation meets learning with noisy labels,'' in
  \emph{ICLR}, 2023.

\bibitem{parnami2022learning}
A.~Parnami and M.~Lee, ``Learning from few examples: A summary of approaches to
  few-shot learning,'' \emph{arXiv:2203.04291}, 2022.

\bibitem{Chi_2022_CVPR}
Z.~Chi, L.~Gu, H.~Liu, Y.~Wang, Y.~Yu, and J.~Tang, ``{MetaFSCIL}: A
  meta-learning approach for few-shot class incremental learning,'' in
  \emph{CVPR}, 2022, pp. 14\,166--14\,175.

\bibitem{survey_fewshot2023}
Y.~Song, T.~Wang, P.~Cai, S.~K. Mondal, and J.~P. Sahoo, ``A comprehensive
  survey of few-shot learning: Evolution, applications, challenges, and
  opportunities,'' \emph{ACM Comput. Surv.}, 2023.

\bibitem{zhao2021domain}
A.~Zhao, M.~Ding, Z.~Lu, T.~Xiang, Y.~Niu, J.~Guan, and J.-R. Wen,
  ``Domain-adaptive few-shot learning,'' in \emph{WACV}, 2021, pp. 1390--1399.

\bibitem{yue2021prototypical}
X.~Yue, Z.~Zheng, S.~Zhang, Y.~Gao, T.~Darrell \emph{et~al.}, ``Prototypical
  cross-domain self-supervised learning for few-shot unsupervised domain
  adaptation,'' in \emph{CVPR}, 2021, pp. 13\,834--13\,844.

\bibitem{hu2022switch}
Z.~Hu, Y.~Sun, and Y.~Yang, ``Switch to generalize: Domain-switch learning for
  cross-domain few-shot classification,'' in \emph{ICLR}, 2022.

\bibitem{pourpanah2022review}
F.~Pourpanah, M.~Abdar, Y.~Luo, X.~Zhou, R.~Wang, C.~P. Lim, X.-Z. Wang, and
  Q.~J. Wu, ``A review of generalized zero-shot learning methods,'' \emph{IEEE
  PAMI}, 2022.

\bibitem{zhou2023comprehensive}
C.~Zhou, Q.~Li, C.~Li, J.~Yu, Y.~Liu, G.~Wang \emph{et~al.}, ``A comprehensive
  survey on pretrained foundation models: A history from {BERT} to {ChatGPT},''
  \emph{arXiv:2302.09419}, 2023.

\bibitem{ijcai2022p762}
Y.~Du, Z.~Liu, J.~Li, and W.~X. Zhao, ``A survey of vision-language pre-trained
  models,'' in \emph{IJCAI}, 2022, pp. 5436--5443.

\bibitem{wang2023internimage}
W.~Wang, J.~Dai, Z.~Chen, Z.~Huang, Z.~Li \emph{et~al.}, ``Internimage:
  Exploring large-scale vision foundation models with deformable
  convolutions,'' in \emph{CVPR}, 2023, pp. 14\,408--14\,419.

\bibitem{baevski2020wav2vec}
A.~Baevski, Y.~Zhou, A.~Mohamed, and M.~Auli, ``wav2vec 2.0: A framework for
  self-supervised learning of speech representations,'' \emph{NeurIPS},
  vol.~33, pp. 12\,449--12\,460, 2020.

\bibitem{tian2021understanding}
Y.~Tian, X.~Chen, and S.~Ganguli, ``Understanding self-supervised learning
  dynamics without contrastive pairs,'' in \emph{ICML}.\hskip 1em plus 0.5em
  minus 0.4em\relax PMLR, 2021, pp. 10\,268--10\,278.

\bibitem{garrido2023on}
Q.~Garrido, Y.~Chen, A.~Bardes, L.~Najman, and Y.~LeCun, ``On the duality
  between contrastive and non-contrastive self-supervised learning,'' in
  \emph{ICLR}, 2023.

\bibitem{baevski2022data2vec}
A.~Baevski, W.-N. Hsu, Q.~Xu, A.~Babu, J.~Gu, and M.~Auli, ``Data2vec: A
  general framework for self-supervised learning in speech, vision and
  language,'' in \emph{ICML}.\hskip 1em plus 0.5em minus 0.4em\relax PMLR,
  2022, pp. 1298--1312.

\bibitem{radford2021learning}
A.~Radford, J.~W. Kim, C.~Hallacy, A.~Ramesh, G.~Goh \emph{et~al.}, ``Learning
  transferable visual models from natural language supervision,'' in
  \emph{ICML}.\hskip 1em plus 0.5em minus 0.4em\relax PMLR, 2021, pp.
  8748--8763.

\bibitem{babu2021xlsr}
A.~Babu, C.~Wang, A.~Tjandra, K.~Lakhotia, Q.~Xu \emph{et~al.}, ``Xls-r:
  Self-supervised cross-lingual speech representation learning at scale,''
  \emph{arXiv}, vol. abs/2111.09296, 2021.

\bibitem{scao2022bloom}
T.~L. Scao, A.~Fan, C.~Akiki, E.~Pavlick, S.~Ili{\'c} \emph{et~al.}, ``Bloom: A
  176b-parameter open-access multilingual language model,''
  \emph{arXiv:2211.05100}, 2022.

\bibitem{ding2022delta}
N.~Ding, Y.~Qin, G.~Yang, F.~Wei, Z.~Yang \emph{et~al.}, ``Delta tuning: A
  comprehensive study of parameter efficient methods for pre-trained language
  models,'' \emph{arXiv:2203.06904}, 2022.

\bibitem{pmlr_houlsby19a}
N.~Houlsby, A.~Giurgiu, S.~Jastrzebski, B.~Morrone, Q.~De~Laroussilhe
  \emph{et~al.}, ``Parameter-efficient transfer learning for {NLP},'' in
  \emph{ICML}, vol.~97.\hskip 1em plus 0.5em minus 0.4em\relax PMLR, 2019, pp.
  2790--2799.

\bibitem{li-liang-2021-prefix}
X.~L. Li and P.~Liang, ``Prefix-tuning: Optimizing continuous prompts for
  generation,'' in \emph{IJCNLP-AACL}, 2021, pp. 4582--4597.

\bibitem{lester2021power}
B.~Lester, R.~Al-Rfou, and N.~Constant, ``The power of scale for
  parameter-efficient prompt tuning,'' in \emph{Proceedings of EMNLP}, 2021,
  pp. 3045--3059.

\bibitem{hu2021lora}
E.~J. Hu, yelong shen, P.~Wallis, Z.~Allen-Zhu, Y.~Li, S.~Wang, L.~Wang, and
  W.~Chen, ``Lora: Low-rank adaptation of large language models,'' in
  \emph{ICLR}, 2022.

\bibitem{he2022towards}
J.~He, C.~Zhou, X.~Ma, T.~Berg-Kirkpatrick, and G.~Neubig, ``Towards a unified
  view of parameter-efficient transfer learning,'' in \emph{ICLR}, 2022.

\bibitem{sun19ttt}
Y.~Sun, X.~Wang, L.~Zhuang, J.~Miller, M.~Hardt, and A.~A. Efros, ``Test-time
  training with self-supervision for generalization under distribution
  shifts,'' in \emph{ICML}, 2020.

\bibitem{Chi_2021_CVPR}
Z.~Chi, Y.~Wang, Y.~Yu, and J.~Tang, ``Test-time fast adaptation for dynamic
  scene deblurring via meta-auxiliary learning,'' in \emph{CVPR}, 2021, pp.
  9137--9146.

\bibitem{goyaltest}
S.~Goyal, M.~Sun, A.~Raghunathan, and J.~Z. Kolter, ``Test time adaptation via
  conjugate pseudo-labels,'' in \emph{NeurIPS}, 2022.

\bibitem{chen2022contrastive}
D.~Chen, D.~Wang, T.~Darrell, and S.~Ebrahimi, ``Contrastive test-time
  adaptation,'' in \emph{CVPR}, 2022.

\bibitem{NEURIPS2021_b618c321}
Y.~Liu, P.~Kothari, B.~van Delft, B.~Bellot-Gurlet, T.~Mordan, and A.~Alahi,
  ``Ttt++: When does self-supervised test-time training fail or thrive?'' in
  \emph{NeurIPS}, vol.~34.\hskip 1em plus 0.5em minus 0.4em\relax Curran
  Associates, Inc., 2021, pp. 21\,808--21\,820.

\bibitem{gandelsmantest}
Y.~Gandelsman, Y.~Sun, X.~Chen, and A.~A. Efros, ``Test-time training with
  masked autoencoders,'' in \emph{NeurIPS}, 2022.

\bibitem{su2022revisiting}
Y.~Su, X.~Xu, and K.~Jia, ``Revisiting realistic test-time training: Sequential
  inference and adaptation by anchored clustering,'' in \emph{NeurIPS}, 2022.

\bibitem{iwasawa2021test}
Y.~Iwasawa and Y.~Matsuo, ``Test-time classifier adjustment module for
  model-agnostic domain generalization,'' \emph{NeurIPS}, vol.~34, pp.
  2427--2440, 2021.

\bibitem{niu2022efficient}
S.~Niu, J.~Wu, Y.~Zhang, Y.~Chen, S.~Zheng, P.~Zhao, and M.~Tan, ``Efficient
  test-time model adaptation without forgetting,'' in \emph{ICML}.\hskip 1em
  plus 0.5em minus 0.4em\relax PMLR, 2022, pp. 16\,888--16\,905.

\bibitem{wang2022continual}
Q.~Wang, O.~Fink, L.~Van~Gool, and D.~Dai, ``Continual test-time domain
  adaptation,'' in \emph{CVPR}, 2022, pp. 7201--7211.

\bibitem{sogaard-etal-2021-need}
A.~S{\o}gaard, S.~Ebert, J.~Bastings, and K.~Filippova, ``We need to talk about
  random splits,'' in \emph{EACL}, 2021, pp. 1823--1832.

\bibitem{gama2004learning}
J.~Gama, P.~Medas, G.~Castillo, and P.~Rodrigues, ``Learning with drift
  detection,'' in \emph{Advances in Artificial Intelligence--SBIA}.\hskip 1em
  plus 0.5em minus 0.4em\relax Springer, 2004, pp. 286--295.

\bibitem{bifet2007learning}
A.~Bifet and R.~Gavalda, ``Learning from time-changing data with adaptive
  windowing,'' in \emph{SDM}.\hskip 1em plus 0.5em minus 0.4em\relax SIAM,
  2007, pp. 443--448.

\bibitem{9729470}
A.~Liu, J.~Lu, Y.~Song, J.~Xuan, and G.~Zhang, ``Concept drift detection delay
  index,'' \emph{IEEE Transactions on Knowledge and Data Engineering}, pp.
  1--1, 2022.

\bibitem{yu2018request}
S.~Yu, X.~Wang, and J.~C. Pr{\'\i}ncipe, ``Request-and-reverify: hierarchical
  hypothesis testing for concept drift detection with expensive labels,'' in
  \emph{IJCAI}, 2018, pp. 3033--3039.

\bibitem{sidhu2015online}
P.~Sidhu and M.~Bhatia, ``An online ensembles approach for handling concept
  drift in data streams: diversified online ensembles detection,''
  \emph{International Journal of Machine Learning and Cybernetics}, vol.~6,
  no.~6, pp. 883--909, 2015.

\bibitem{khamassi2019new}
I.~Khamassi, M.~Sayed-Mouchaweh, M.~Hammami, and K.~Gh{\'e}dira, ``A new
  combination of diversity techniques in ensemble classifiers for handling
  complex concept drift,'' \emph{Learning from Data Streams in Evolving
  Environments: Methods and Applications}, pp. 39--61, 2019.

\bibitem{anderson2019recurring}
R.~Anderson, Y.~S. Koh, G.~Dobbie, and A.~Bifet, ``Recurring concept
  meta-learning for evolving data streams,'' \emph{Expert Systems with
  Applications}, vol. 138, p. 112832, 2019.

\bibitem{YU2022996}
H.~Yu, Q.~Zhang, T.~Liu, J.~Lu, Y.~Wen, and G.~Zhang, ``Meta-add: A
  meta-learning based pre-trained model for concept drift active detection,''
  \emph{Information Sciences}, vol. 608, pp. 996--1009, 2022.

\bibitem{zubarouglu2021data}
A.~Zubaro{\u{g}}lu and V.~Atalay, ``Data stream clustering: a review,''
  \emph{Artificial Intelligence Review}, vol.~54, no.~2, pp. 1201--1236, 2021.

\bibitem{oliveira2021tackling}
G.~Oliveira, L.~L. Minku, and A.~L. Oliveira, ``Tackling virtual and real
  concept drifts: An adaptive gaussian mixture model approach,'' \emph{IEEE
  Transactions on Knowledge and Data Engineering}, 2021.

\bibitem{liu2018accumulating}
A.~Liu, J.~Lu, F.~Liu, and G.~Zhang, ``Accumulating regional density
  dissimilarity for concept drift detection in data streams,'' \emph{Pattern
  Recognition}, vol.~76, pp. 256--272, 2018.

\bibitem{castellani2021task}
A.~Castellani, S.~Schmitt, and B.~Hammer, ``Task-sensitive concept drift
  detector with constraint embedding,'' in \emph{SSCI}.\hskip 1em plus 0.5em
  minus 0.4em\relax IEEE, 2021, pp. 01--08.

\bibitem{li2022ddg}
W.~Li, X.~Yang, W.~Liu, Y.~Xia, and J.~Bian, ``{DDG-DA}: Data distribution
  generation for predictable concept drift adaptation,'' in \emph{AAAI},
  vol.~36, no.~4, 2022, pp. 4092--4100.

\bibitem{RAAB2020340}
C.~Raab, M.~Heusinger, and F.-M. Schleif, ``Reactive soft prototype computing
  for concept drift streams,'' \emph{Neurocomputing}, vol. 416, pp. 340--351,
  2020.

\bibitem{9748034}
P.~Li, Y.~Liu, Y.~Hu, Y.~Zhang, X.~Hu, and K.~Yu, ``A drift-sensitive
  distributed lstm method for short text stream classification,'' \emph{IEEE
  Transactions on Big Data}, vol.~9, no.~1, pp. 341--357, 2023.

\bibitem{9047166}
A.~Liu, J.~Lu, and G.~Zhang, ``Diverse instance-weighting ensemble based on
  region drift disagreement for concept drift adaptation,'' \emph{IEEE Trans.
  NNLS}, vol.~32, no.~1, pp. 293--307, 2021.

\bibitem{ghomeshi2019eacd}
H.~Ghomeshi, M.~M. Gaber, and Y.~Kovalchuk, ``Eacd: evolutionary adaptation to
  concept drifts in data streams,'' \emph{Data Mining and Knowledge Discovery},
  vol.~33, pp. 663--694, 2019.

\bibitem{heusinger2022passive}
M.~Heusinger, C.~Raab, and F.-M. Schleif, ``Passive concept drift handling via
  variations of learning vector quantization,'' \emph{Neural Computing and
  Applications}, vol.~34, no.~1, pp. 89--100, 2022.

\bibitem{9783029}
W.~Zheng, P.~Zhao, G.~Chen, H.~Zhou, and Y.~Tian, ``A hybrid spiking neurons
  embedded lstm network for multivariate time series learning under
  concept-drift environment,'' \emph{IEEE Transactions on Knowledge and Data
  Engineering}, pp. 1--1, 2022.

\bibitem{chalkidis2022improved}
I.~Chalkidis and A.~S{\o}gaard, ``Improved multi-label classification under
  temporal concept drift: Rethinking group-robust algorithms in a label-wise
  setting,'' in \emph{ACL}, 2022, pp. 2441--2454.

\bibitem{li2020incremental}
Z.~Li, W.~Huang, Y.~Xiong, S.~Ren, and T.~Zhu, ``Incremental learning
  imbalanced data streams with concept drift: The dynamic updated ensemble
  algorithm,'' \emph{Knowledge-Based Systems}, vol. 195, p. 105694, 2020.

\bibitem{jung2015modeling}
H.~J. Jung and M.~Lease, ``Modeling temporal crowd work quality with limited
  supervision,'' in \emph{AAAI}, vol.~3, 2015, pp. 83--91.

\bibitem{SIGIR2016}
I.~Abraham, O.~Alonso, V.~Kandylas, R.~Patel, S.~Shelford, and A.~Slivkins,
  ``How many workers to ask? adaptive exploration for collecting high quality
  labels,'' in \emph{SIGIR}.\hskip 1em plus 0.5em minus 0.4em\relax Association
  for Computing Machinery, 2016, p. 473–482.

\bibitem{Goyal_McDonnell_Kutlu_Elsayed_Lease_2018}
T.~Goyal, T.~McDonnell, M.~Kutlu, T.~Elsayed, and M.~Lease, ``Your behavior
  signals your reliability: Modeling crowd behavioral traces to ensure quality
  relevance annotations,'' \emph{AAAI}, vol.~6, no.~1, pp. 41--49, 2018.

\bibitem{zafari2019modelling}
F.~Zafari, I.~Moser, and T.~Baarslag, ``Modelling and analysis of temporal
  preference drifts using a component-based factorised latent approach,''
  \emph{Expert Systems with Applications}, vol. 116, pp. 186--208, 2019.

\bibitem{8657692}
Y.~Xie, J.~Li, T.~Zhu, and C.~Liu, ``Continuous-valued annotations aggregation
  for heart rate detection,'' \emph{IEEE Access}, vol.~7, pp. 37\,664--37\,671,
  2019.

\bibitem{9492291}
S.~Liu, S.~Xue, J.~Wu, C.~Zhou, J.~Yang, Z.~Li, and J.~Cao, ``Online active
  learning for drifting data streams,'' \emph{IEEE Trans. NNLS}, vol.~34,
  no.~1, pp. 186--200, 2023.

\bibitem{zhan2022comparative}
X.~Zhan, Q.~Wang, K.-h. Huang, H.~Xiong, D.~Dou, and A.~B. Chan, ``A
  comparative survey of deep active learning,'' \emph{arXiv:2203.13450}, 2022.

\bibitem{yu2022cold}
Y.~Yu, R.~Zhang, R.~Xu, J.~Zhang, J.~Shen, and C.~Zhang, ``Cold-start data
  selection for few-shot language model fine-tuning: A prompt-based uncertainty
  propagation approach,'' \emph{arXiv:2209.06995}, 2022.

\bibitem{hacohen2022active}
G.~Hacohen, A.~Dekel, and D.~Weinshall, ``Active learning on a budget: Opposite
  strategies suit high and low budgets,'' in \emph{ICML}.\hskip 1em plus 0.5em
  minus 0.4em\relax PMLR, 2022, pp. 8175--8195.

\bibitem{korycki2021streaming}
{\L}.~Korycki and B.~Krawczyk, ``Streaming decision trees for lifelong
  learning,'' in \emph{ECML PKDD}.\hskip 1em plus 0.5em minus 0.4em\relax
  Springer, 2021, pp. 502--518.

\bibitem{pratama2019automatic}
M.~Pratama, C.~Za'in, A.~Ashfahani, Y.~S. Ong, and W.~Ding, ``Automatic
  construction of multi-layer perceptron network from streaming examples,'' in
  \emph{CIKM}, 2019, pp. 1171--1180.

\bibitem{LIU2016322}
D.~Liu, Y.~Wu, and H.~Jiang, ``Fp-elm: An online sequential learning algorithm
  for dealing with concept drift,'' \emph{Neurocomputing}, vol. 207, pp.
  322--334, 2016.

\bibitem{van2022three}
G.~M. van~de Ven, T.~Tuytelaars, and A.~S. Tolias, ``Three types of incremental
  learning,'' \emph{Nature Machine Intelligence}, pp. 1--13, 2022.

\bibitem{Korycki_2021_CVPR}
L.~Korycki and B.~Krawczyk, ``Class-incremental experience replay for continual
  learning under concept drift,'' in \emph{CVPR Workshops}, June 2021, pp.
  3649--3658.

\bibitem{darem2021adaptive}
A.~A. Darem, F.~A. Ghaleb, A.~A. Al-Hashmi, J.~H. Abawajy, S.~M. Alanazi, and
  A.~Y. Al-Rezami, ``An adaptive behavioral-based incremental batch learning
  malware variants detection model using concept drift detection and sequential
  deep learning,'' \emph{IEEE Access}, vol.~9, pp. 97\,180--97\,196, 2021.

\bibitem{Taufique_2022_CVPR}
A.~M.~N. Taufique, C.~S. Jahan, and A.~Savakis, ``Unsupervised continual
  learning for gradually varying domains,'' in \emph{CVPR Workshops}, 2022, pp.
  3740--3750.

\bibitem{9359505}
J.~Li, M.~Jing, H.~Su, K.~Lu, L.~Zhu, and H.~T. Shen, ``Faster domain
  adaptation networks,'' \emph{IEEE Transactions on Knowledge and Data
  Engineering}, vol.~34, no.~12, pp. 5770--5783, 2022.

\bibitem{9802893}
B.~Jiao, Y.~Guo, D.~Gong, and Q.~Chen, ``Dynamic ensemble selection for
  imbalanced data streams with concept drift,'' \emph{IEEE Trans. NNLS}, pp.
  1--14, 2022.

\bibitem{9961859}
B.~Jiao, Y.~Guo, C.~Yang, J.~Pu, Z.~Zheng, and D.~Gong, ``Incremental weighted
  ensemble for data streams with concept drift,'' \emph{IEEE Transactions on
  Artificial Intelligence}, pp. 1--12, 2022.

\bibitem{kuo2022green}
C.-C.~J. Kuo and A.~M. Madni, ``Green learning: Introduction, examples and
  outlook,'' \emph{Journal of Visual Communication and Image Representation},
  p. 103685, 2022.

\bibitem{HUANG2020100270}
X.~Huang, D.~Kroening, W.~Ruan, J.~Sharp \emph{et~al.}, ``A survey of safety
  and trustworthiness of deep neural networks: Verification, testing,
  adversarial attack and defence, and interpretability,'' \emph{Computer
  Science Review}, vol.~37, p. 100270, 2020.

\bibitem{ijcai2021p591}
T.~Bai, J.~Luo, J.~Zhao, B.~Wen, and Q.~Wang, ``Recent advances in adversarial
  training for adversarial robustness,'' in \emph{IJCAI}, 2021, pp. 4312--4321.

\bibitem{wang2023robustness}
J.~Wang, H.~Xixu, W.~Hou, H.~Chen, R.~Zheng, Y.~Wang \emph{et~al.}, ``On the
  robustness of chatgpt: An adversarial and out-of-distribution perspective,''
  in \emph{ICLR Workshop}, 2023.

\bibitem{jin2021differentially}
K.~Jin, X.~Cheng, J.~Yang, and K.~Shen, ``Differentially private correlation
  alignment for domain adaptation.'' in \emph{IJCAI}, vol.~21, 2021, pp.
  3649--3655.

\bibitem{peterson2019private}
D.~Peterson, P.~Kanani, and V.~J. Marathe, ``Private federated learning with
  domain adaptation,'' \emph{arXiv:1912.06733}, 2019.

\bibitem{9685083}
D.~M. Manias, I.~Shaer, L.~Yang, and A.~Shami, ``Concept drift detection in
  federated networked systems,'' in \emph{GLOBECOM}, 2021, pp. 1--6.

\bibitem{casado2022concept}
F.~E. Casado, D.~Lema, M.~F. Criado, R.~Iglesias, C.~V. Regueiro, and S.~Barro,
  ``Concept drift detection and adaptation for federated and continual
  learning,'' \emph{Multimedia Tools and Applications}, pp. 1--23, 2022.

\bibitem{mukherjee2022domain}
D.~Mukherjee, F.~Petersen, M.~Yurochkin, and Y.~Sun, ``Domain adaptation meets
  individual fairness. and they get along.'' in \emph{NeurIPS}, 2022.

\bibitem{wang2022robust}
H.~Wang, J.~Hong, J.~Zhou, and Z.~Wang, ``How robust is your fairness?
  evaluating and sustaining fairness under unseen distribution shifts,''
  \emph{arXiv:2207.01168}, 2022.

\bibitem{madras2018learning}
D.~Madras, E.~Creager, T.~Pitassi, and R.~Zemel, ``Learning adversarially fair
  and transferable representations,'' in \emph{ICML}.\hskip 1em plus 0.5em
  minus 0.4em\relax PMLR, 2018, pp. 3384--3393.

\bibitem{ZENISEK2021507}
J.~Zenisek, N.~Wild, and J.~Wolfartsberger, ``Investigating the potential of
  smart manufacturing technologies,'' \emph{Procedia Computer Science}, vol.
  180, pp. 507--516, 2021.

\bibitem{survey_drift2014}
J.~a. Gama, I.~\v{Z}liobaitundefined, A.~Bifet, M.~Pechenizkiy, and
  A.~Bouchachia, ``A survey on concept drift adaptation,'' \emph{ACM Comput.
  Surv.}, vol.~46, no.~4, 2014.

\bibitem{lin2019concept}
C.-C. Lin, D.-J. Deng, C.-H. Kuo, and L.~Chen, ``Concept drift detection and
  adaption in big imbalance industrial iot data using an ensemble learning
  method of offline classifiers,'' \emph{IEEE Access}, vol.~7, pp.
  56\,198--56\,207, 2019.

\bibitem{thiagarajan2019understanding}
J.~J. Thiagarajan, D.~Rajan, and P.~Sattigeri, ``Understanding behavior of
  clinical models under domain shifts,'' in \emph{ACM SIGKDD}, 2019.

\bibitem{9674886}
L.~Ju, X.~Wang, L.~Wang, D.~Mahapatra, X.~Zhao, Q.~Zhou, T.~Liu, and Z.~Ge,
  ``Improving medical images classification with label noise using
  dual-uncertainty estimation,'' \emph{IEEE Transactions on Medical Imaging},
  vol.~41, no.~6, pp. 1533--1546, 2022.

\end{thebibliography}


 




\begin{IEEEbiography}[{\includegraphics[width=1in,height=1.25in,clip,keepaspectratio]{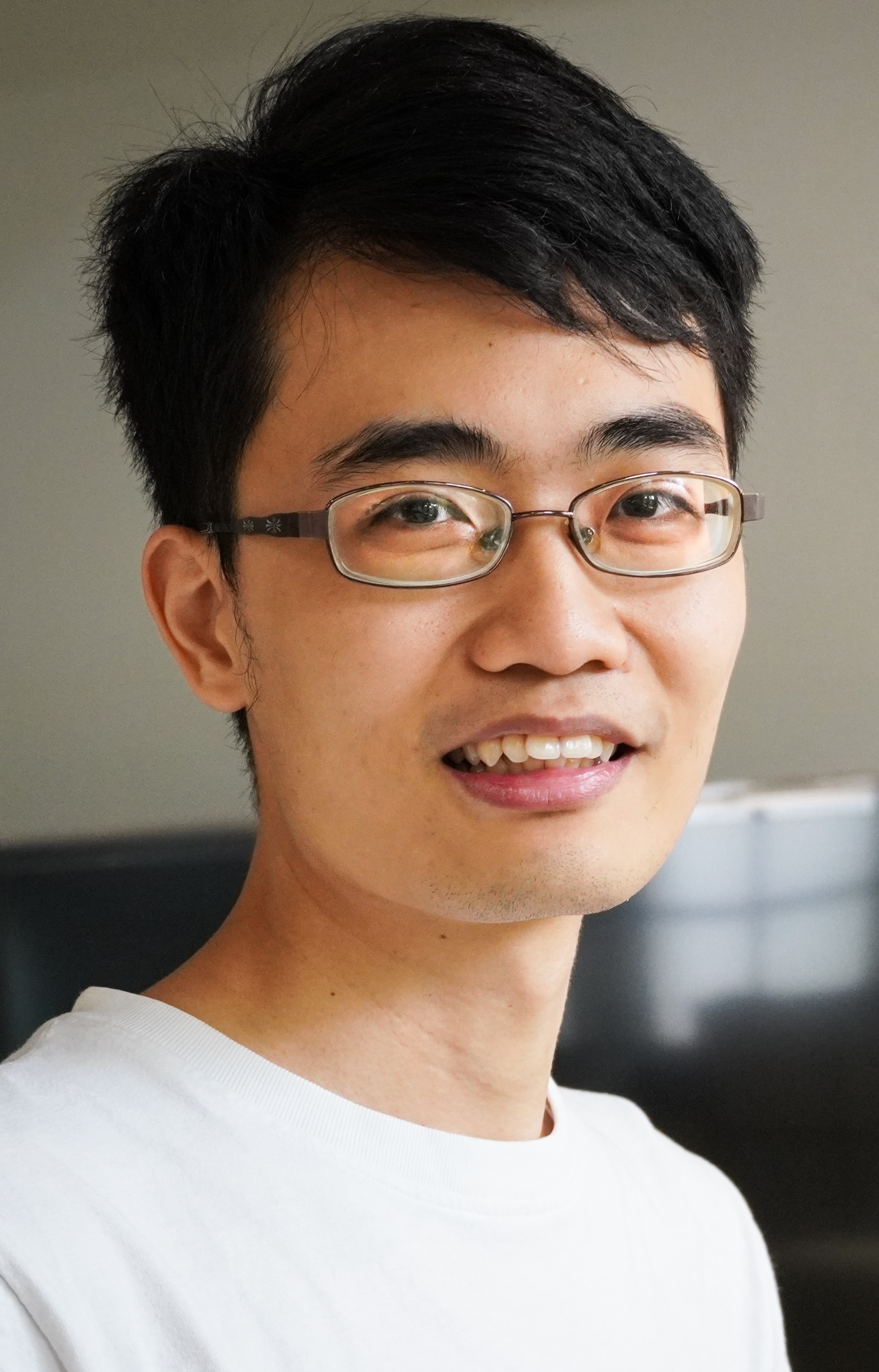}}]{Jeng-Lin Li}
is currently a Senior Data Scientist at Inventec Corporation and an Adjunct Assistant Professor in the Biomed AI PhD program at National Tsing Hua University (NTHU). He received his B.S. (2016) and Ph.D. (2022) degrees from the Department of Electrical Engineering at NTHU. He was awarded ACLCLP dissertation award (2022), NTHU Principal Outstanding Student Scholarship (2017-2020), industrial scholarships from Garmin (2018), Yahoo (2019), Novatek (2020, 2021), and travel grants from EMBC (2019), Interspeech (2019). His research interests are machine learning, health analytics, behavior signal processing, and affective computing. 
\end{IEEEbiography}

\begin{IEEEbiography}
[{\includegraphics[width=1in,height=1.25in,clip,keepaspectratio]{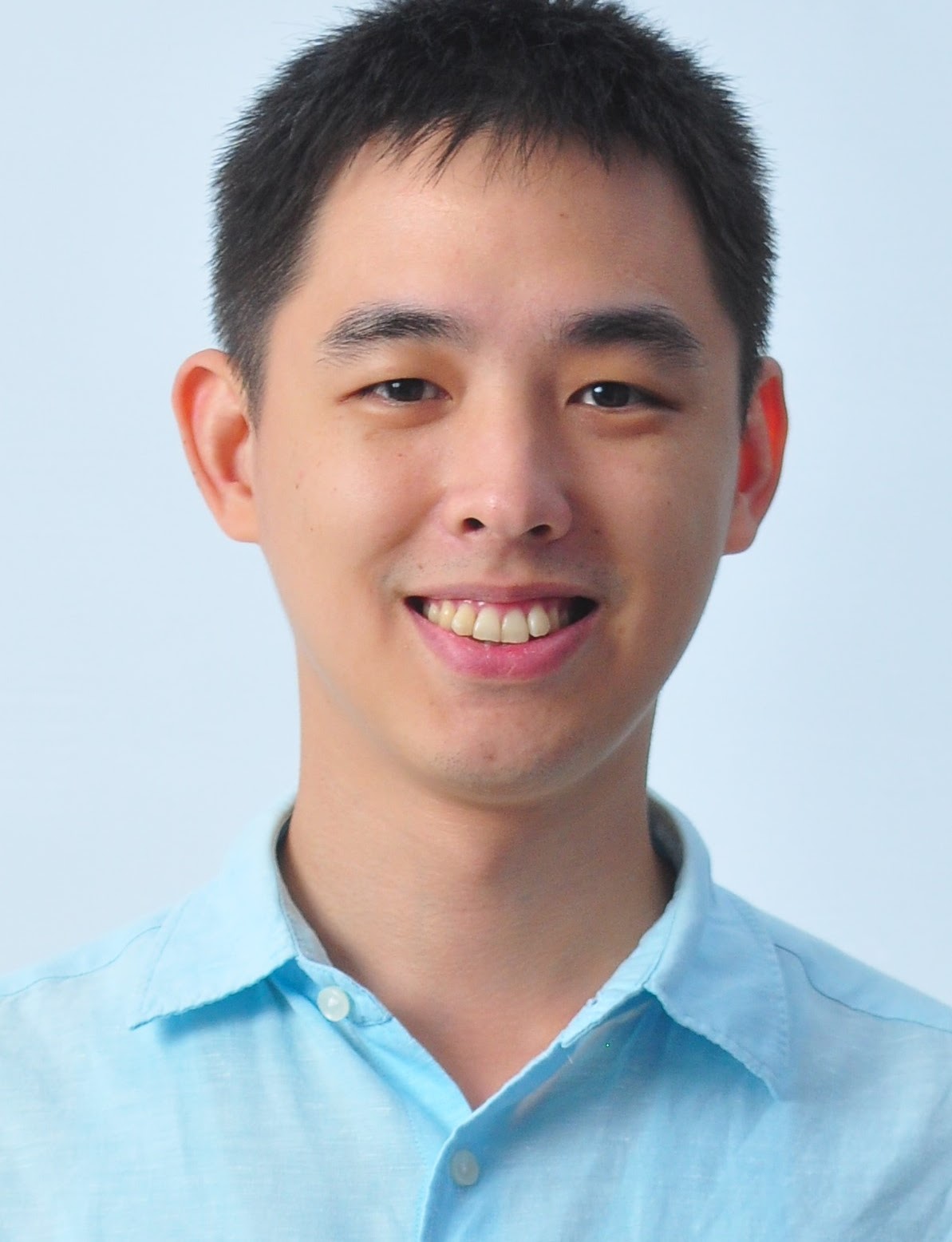}}]{Chih-Fan Hsu}
Chih-Fan Hsu is a Senior Data Scientist at Inventec Corporation. His research interests include computer vision, machine learning, virtual reality, and multimedia systems. Dr. Hsu was a Postdoc at the University of California, Davis, National Tsing Hua University, and National Yang Ming Chiao Tung University (2020-2021), and a research assistant at the Academia Sinica (2014-2020). Dr. Hsu received his MS in Computer Science and Information Engineering from the National Taiwan Normal University (2010) and Ph.D. in Electrical Engineering from the National Taiwan University (2019).
\end{IEEEbiography}

\begin{IEEEbiography}[{\includegraphics[width=1in,height=1.25in,clip,keepaspectratio]{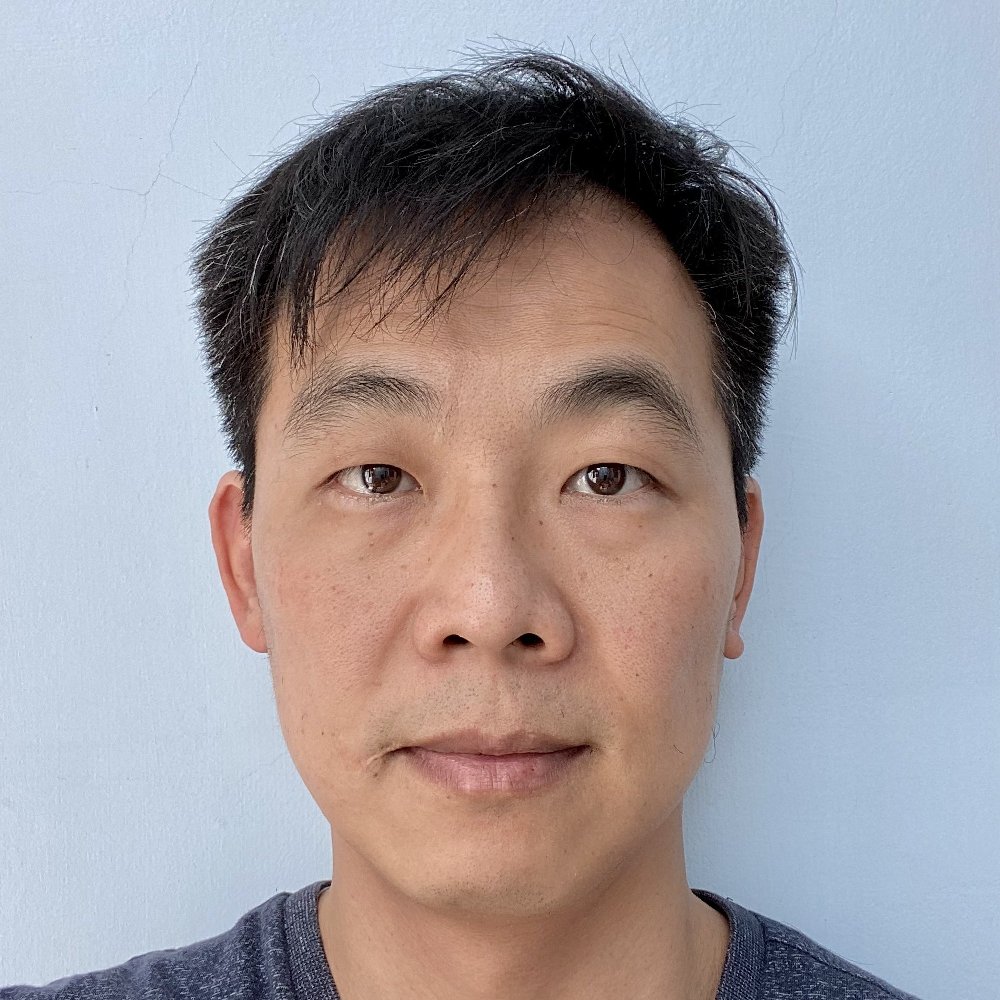}}]{Ming-Ching Chang}
is an Associate Professor at the Department of Computer Science, University at Albany, SUNY. His expertise includes AI, video analytics, computer vision, and machine learning. His research projects are funded by DARPA, IARPA, NIJ, VA, GE Global Research. 
Dr. Chang frequently serves the program chair, area chair, and referee of leading journals and conferences. 
He chairs the steering committee of the IEEE AVSS Conference since 2022, which he serves as the committee member since 2017. 
He has authored more than 100 peer-reviewed journal and conference publications, 7 US patents and 15 disclosures. He is a senior member of IEEE and member of ACM.
\end{IEEEbiography}

\begin{IEEEbiography}[{\includegraphics[width=1in,height=1.25in,clip,keepaspectratio]{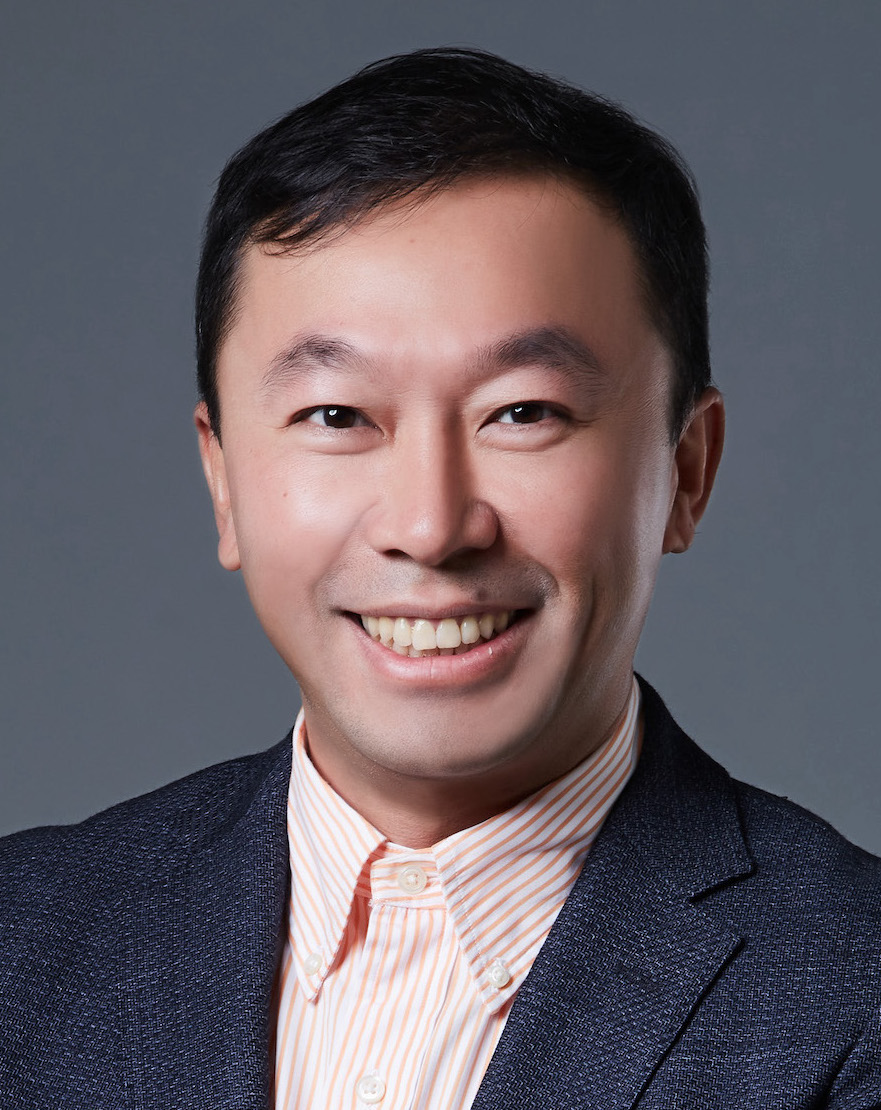}}]{Wei-Chao Chen} is the Chief Digital Officer at Inventec Corporation and the Chairman at Skywatch Inc.  Dr. Chen is also a Visiting Professor at the National Taiwan University.  His research interests include graphics hardware, computational photography, augmented reality, and computer vision.  Dr. Chen was the Chief AI Advisor at Inventec (2018-2020), a senior research scientist in Nokia Research Center at Palo Alto (2007-2009), and a 3D Graphics Architect in NVIDIA (2002-2006).  Dr. Chen received his MS in Electrical Engineering from National Taiwan University (1996) and Ph.D. in Computer Science from the University of North Carolina at Chapel Hill (2002).
\end{IEEEbiography}

\vfill

\end{document}